\documentclass[letterpaper, 10pt]{IEEEtran}
\IEEEoverridecommandlockouts
\usepackage[pdftex]{graphicx}

\graphicspath{{../pdf/}{../jpeg/}}
\DeclareGraphicsExtensions{.pdf,.jpeg,.png,.jpg}

\usepackage{amssymb} 
\usepackage{graphicx}
\usepackage[natural]{xcolor}

\usepackage{changes}

\usepackage{siunitx}

\usepackage{caption}

\usepackage{amsmath}
\usepackage[tight,footnotesize]{subfigure}

\usepackage{url}

\usepackage{colortbl}
\usepackage{tabulary}
\usepackage{etoolbox}
\usepackage{tabularx}
\usepackage{multirow}
\usepackage{float}
\usepackage{bm}
\usepackage{booktabs}

\definecolor{a}{HTML}{53CF00}
\definecolor{b}{HTML}{C1D900}
\definecolor{c}{HTML}{C2AC00}
\definecolor{d}{HTML}{B58700}
\definecolor{e}{HTML}{ED8F00}
\definecolor{f}{HTML}{B54000}
\definecolor{g}{HTML}{ED2300}

\begin{document}

\title{\LARGE{ \bf Extended Coopetitive Soft Gating Ensemble}} 


\author{Stephan Deist, Jens Schreiber, Maarten Bieshaar and Bernhard Sick\\
	  Intelligent Embedded Systems Group, University of Kassel (Germany)\\	
		email: \{stephan.deist, jens.schreiber, mbieshaar,  bsick\}@uni-kassel.de
	%
}

\maketitle	
\begin{abstract}
This article is about an extension of a recent \textit{ensemble method} called \textit{Coopetitive Soft Gating Ensemble (CSGE)} and its application on power forecasting as well as motion primitive forecasting of cyclists.
The \textit{CSGE} has been used successfully in the field of wind power forecasting, outperforming common algorithms in this domain.
The principal idea of the \textit{CSGE} is to weight the models regarding their observed performance during training on different aspects.
Several extensions are proposed to the original \textit{CSGE} within this article, making the \textit{ensemble} even more flexible and powerful.
The extended \textit{CSGE} (\textit{XCSGE} as we term it), is used to predict the power generation on both wind- and solar farms. Moreover, the \textit{XCSGE} is applied to forecast the movement state of cyclists in the context of driver assistance systems.
Both domains have different requirements, are non-trivial problems, and are used to evaluate various facets of the novel \textit{XCSGE}.
The two problems differ fundamentally in the size of the data sets and the number of features. Power forecasting is based on weather forecasts that are subject to fluctuations in their features. In the movement primitive forecasting of cyclists, time delays contribute to the difficulty of the prediction.
The \textit{XCSGE} reaches an improvement of the prediction performance of up to 11\% for wind power forecasting and 30\% for solar power forecasting compared to the worst performing model.
For the classification of movement primitives of cyclists, the \textit{XCSGE} reaches an improvement of up to 28\%.
The evaluation includes a comparison with other state-of-the-art \textit{ensemble methods}. We can verify that the \textit{XCSGE} results are significantly better using the Nemenyi post-hoc test. 
\end{abstract}



\section{Introduction}
\label{sec:einleitung}

\begin{figure}
  \centering
    \includegraphics[scale=0.3]{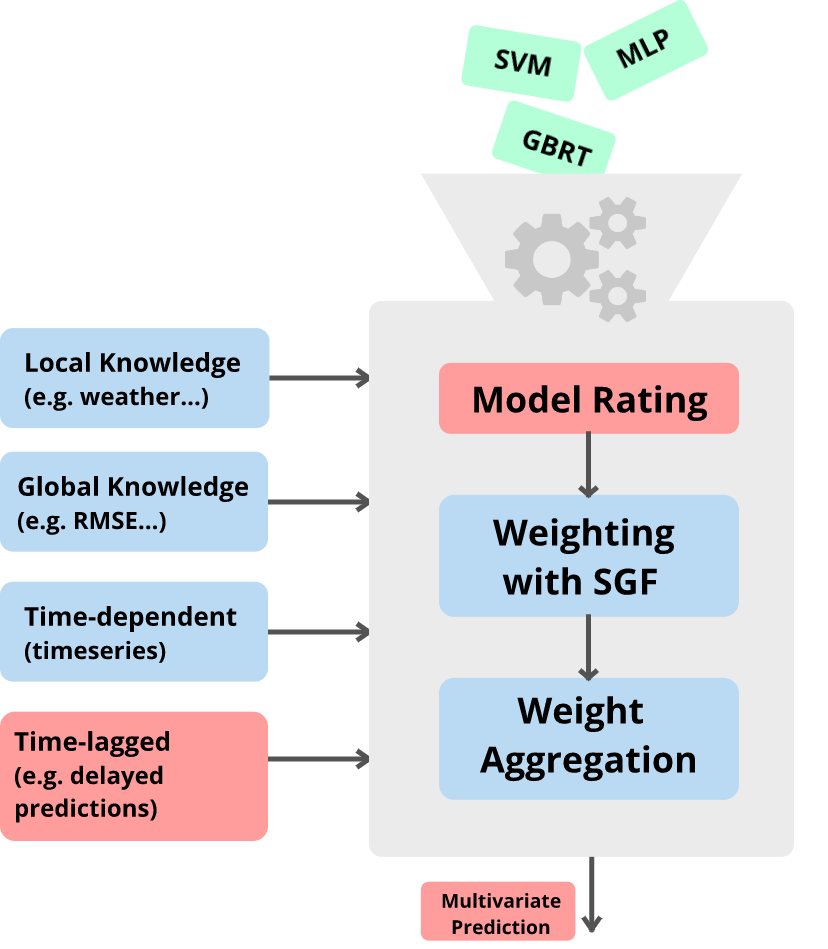}
\vspace{10px}
     \caption{\small Aspects that the \textit{XCSGE} includes in the weighting. In green the individual \textit{ensemble members} are marked.The methodological extentions of the \textit{XCSGE} presented within this article are highlighted in red. The extensions mainly concern the model evaluation, time lagged and the possibility for multivariate predictions. \label{ensemble_simple}}
\end{figure}

The main goal of machine learning (ML) is to create models from a set of training data that have a high capability of generalization.
Often the ML problems are so complex that one single model cannot handle the whole scope. A common approach is to have multiple prediction models instead of using only one single model. The approach of combining multiple estimators is called \textit{ensemble method} or short \textit{ensemble}.
In \cite{hansen_salamon} and \cite{zhou}, it is shown that ensembles often lead to better results than using one single estimator.
Despite their good generalization performance, well-known \textit{ensemble methods} such as \textit{boosting} \cite{Schapire1990}, \textit{bagging} \cite{Breiman1996} or \textit{stacking} \cite{Smyth1999} have the disadvantage that the aggregation function is not human-readable. They can be considered as black box models.
This article deals with a novel \textit{ensemble method} called \textit{CSGE}. 
The proposed \textit{ensemble method} attempts to circumvent this decisive disadvantage by weighting the models according to different, potentially influencing aspects that are easy for humans to comprehend.
The different factors can be seen in Figure \ref{ensemble_simple}.
In the following, the example of wind power forecasting is considered.
Several ML models provide predictions that are evaluated by four different factors.
In particular, local factors depending on the current situation, are included in the calculation. This includes, for example, the current weather.
Furthermore, global factors are taken into account, e.g. the error scores of the models on validation data or the position of the wind farm.
In addition, it makes sense to understand various processes as a time series (time-dependent and time-lagged). The mentioned aspects can be found in many task positions of the ML.
The presented ensemble unites all these mentioned aspects in a very understandable way for humans. It evaluates the models according to the mentioned criteria providing a weight aggregation for multivariate predictions.

\subsection{Contribution}

The presented \textit{ensemble} algorithm is mentioned in \cite{gensler_sick_2017}, \cite{gensler_sick_probabilistic}, and \cite{gensler_knn_approx} initially. However, the \textit{CSGE} is strongly designed and evaluated to the needs of wind power forecasting.
In this article, we present a \textbf{general approach} of the \textit{CSGE}, which allows the use for \textbf{regression} as well as \textbf{classification tasks}. In addition, we have extended the \textit{XCSGE} to forecast \textbf{multivariate predictions} and use arbitrary error scores.
Furthermore, we have generalized the concept of \textit{local weighting} and gained the possibility to use an \textbf{arbitrary machine learning method} for this purpose.\\
We investigate the performance of the extended \textit{CSGE} (\textit{XCSGE}) on \textbf{two use-cases}, which include both regression and classification tasks. Both use cases have \textbf{very different requirements} and can be solved efficiently with \textit{ensemble} approaches. Besides the different sizes of the two data sets, the number of features differs strongly (30 or 65 features for wind- and solar power forecasting, respectively 738 features for classification of movement primitives). On the one hand, we use the implemented \textit{XCSGE} to predict the power generation of both wind- and solar farms. In addition, the weather models, which we use as the basis for our predictions, are subject to fluctuations that are taken into account by the \textit{XCSGE}.
On the other hand, we use the \textit{XCSGE} to predict the state of motion of a cyclist using smart devices. The main difficulty lies in a large number of physical features originating from the smart devices sensors (i.e. accelerometer and gyroscope). On both applications, we reach state-of-the-art performance, which we prove with statistical tests. Finally, we also include an evaluation of the models that estimate the expected error for the \textit{local weighting}.

\subsection{Structure}
In the following, we give a short overview of the article by summarizing each chapter. Chapter \ref{sec:related_work} (Related Work) gives a short overview of common \textit{ensemble methods}. Besides, we consider \textit{ensemble methods} that are especially used in the fields of power generation and motion prediction.
In Chapter \ref{sec:method} (Methods), a brief overview of the necessary fundamentals of \textit{ensemble methods} and a detailed description of the \textit{XCSGE} is given.  In addition, the developed extensions, which are included in this article, are explained.
In Chapter \ref{energyforecast} (Renewable Energyforecast), we apply the \textit{XCSGE} to predict the power generation of wind- as well as solar farms. A detailed evaluation of the trained \textit{XCSGE} and a comparison with state-of-the-art \textit{ensemble methods} is given. We further investigate the performance of the \textit{local weighting} models.
In Chapter \ref{cyclists} (Cyclists Basic Movement Detection), we apply the \textit{XCSGE} to predict the motion primitives of cyclists using smart devices. A detailed evaluation of the trained models is given.
The article closes with Chapter \ref{sec:outlook} (Conclusion and Outlook).

\section{Related Work} \label{sec:related_work}
This chapter gives a short overview of common \textit{ensemble methods}. Afterwards we discuss \textit{ensemble methods}, which are specifically used in the field of power forecasting as well as motion prediction.\\
The reason why \textit{ensembles} often achieve better results than individual models is due to their \textit{diversity}. According to \cite{diversity}, \textit{diversity} is a measure of how differently the individual ensemble members spread their errors over the \textit{feature space}. The greater the \textit{diversity} between the individual \textit{ensemble members}, the greater the ensemble's ability to generalize.\\
According to \cite{Ren2016EnsembleCA},  there are several principles to achieve diversity. The most important are \textit{data diversity}, \textit{parameter diversity} as well as \textit{structure diversity}.

\begin{itemize}
\item With the \textit{\textbf{data diversity}} principle, further training sets are generated from the original training set, on which an ensemble member is trained. Well-known examples are \textit{Boosting} \cite{Schapire1990} and \textit{Bagging} \cite{Breiman1996}.
\item The \textit{\textbf{parameter diversity}} principle varies the parameters of the learning algorithm to create different models. A common example is \textit{multiple kernel learning (MKL)} \cite{mkl}.
\item Using the \textit{\textbf{structure diversity}} principle, different learning algorithms are used to train diverse ensemble members. An overview of those can be found in \cite{hetero_ensembles}. 
\end{itemize}

In the following, we will point out popular ensemble approaches which are used in the fields of power forecasting.\\
The authors of \cite{weather_ensemble} give an overview of different \textit{ensemble methods} that are used in the area of power forecasting.
Often the meteorological properties of weather forecasts are used as features to build models on them. Therefore, the approaches are differentiated mainly according to the use of weather forecasts. The three most important approaches are \textit{single-model ensemble}, \textit{multi-model ensemble} and \textit{time-lagged ensemble}.

\begin{itemize}
\item  With the \textit{single-model ensemble} \cite{sme} different weather models are generated by varying hyperparameters (\textit{parameter diversity}), on which then arbitrary power forecasting models are trained.
\item  The \textit{multi-model ensemble} \cite{mme} uses weather models from different providers to train power forecasting models. The \textit{structure diversity} principle (different providers use different learning algorithms) and the \textit{data diversity} principle (different providers measure different physical characteristics) are used here.
\item \textit{time-lagged ensembles} \cite{tle1} usually use a single weather model and an single energy prediction model. \textit{Diversity} comes from a variation of the point in time from which a prediction is made (\textit{data diversity}).
\end{itemize}

In the following, a survey of works is given, which deals with the motion detection of cyclists, respectively, humans. An overview of state-of-the-art techniques in the field of human activity recognition based on wearable sensors is given in \cite{Jordao2018HumanAR}. In \cite{CATAL20151018} an ensemble approach to predict physical activities such as sitting, running, etc. is presented. First various models (\textit{structure diversity}) like an \textit{MLP} \cite{159058} and a \textit{decision tree} \cite{97458} are trained. These models are then combined using a \textit{voting classifier} \cite{RUTA200563}. The results strongly suggest researchers applying an ensemble of classifiers approach for activity recognition problems.\\
In \cite{8754785784} different ensemble approaches such as \textit{boosting}, \textit{bagging} and \textit{ensembles of nested dichotomies (END)} \cite{Frank:2004:END:1015330.1015363} are used to predict six everyday activities using smartphone sensor data. The activity classes are: walking, walking upstairs, walking downstairs, sitting, standing, and lying. Accuracy rates of up to 99.22\% are achieved. In \cite{1707361}, multiple \textit{Kalman filters} are used to determine the position of cyclists using GPS data. For this purpose, the \textit{MMAE} (multiple models adaptive estimation) algorithm is used, which allows several \textit{Kalman filters} to run in parallel using different stochastic models (\textit{parameter diversity}).\\ In \cite{mb_1} the authors use a \textit{Stacking ensemble} to classify the state of motion of cyclists. This ensemble implements the \textit{structure diversity} principle. On the one hand, a \textit{convolutional neural network} (CNN) \cite{cnn} is trained on the basis of camera images. On the other hand, a classifier is being trained on smart device based sensor data. These two classifiers serve as a basis to train a \textit{stacking ensemble}, which uses an \textit{extreme gradient boosting} classifier \cite{friedman2001} as \textit{meta-learner}.\\
In \cite{depping}, the state of motion of cyclists is predicted with the help of smart devices. For this purpose, an \textit{XGBoost (extreme gradient regression trees)} \cite{xgboost} classificator is trained for each smart device (\textit{data diversity}). These three classifiers are then used to train another \textit{XGBoost} classificator as a meta-learner. On the one hand, the meta-learner is being trained on the outputs of the three smart device classifiers. On the other hand, the meta-learner is being trained on the outputs of the three smart devices and the fusioned \textit{feature space}.

\section{Method}
\label{sec:method}
In this chapter, we first give a brief overview of basic concepts in ensemble methods, followed by an introduction to the \textit{XCSGE}. After that, a short overview of the three weighting aspects that the \textit{XCSGE} takes into consideration is given.
Based on this description, we introduce the \textit{Soft Gating Principle}, the key concept of the \textit{XCSGE}, which computes a weighting from the estimated error.
Afterward, the three mentioned weighting aspects are explained in detail.
The chapter closes with an overview of the training process of the \textit{XCSGE}.

\subsection{Ensemble Methods}
\label{sec:ensemble_methods}

The approach of combining multiple estimators is called \textit{ensemble}. 
Each estimator of an \textit{ensemble} is called \textit{ensemble member}, \textit{base estimator} or \textit{base learner}. Most ensemble methods use one single type of machine learning algorithm for their ensemble members. These 
ensembles are often called \textit{homogenous ensembles}.
Some ensembles use different machine learning approaches as ensemble members which leads to \textit{heterogeneous ensembles}.\\ According to \cite{zhou}, ensembles can be categorized in one of three types: \textit{combining classifiers}, \textit{mixture of experts} and \textit{ensembles of weak learners}. In this paper, we mainly consider \textit{combining classifiers or regression models}.
When creating an ensemble of \textit{combining classifiers}, multiple \textit{strong learners} are combined to improve the performance. \textit{Strong learners} are estimators that could also work on their own and have an acceptable performance. In contrast, \textit{weak learners} are estimators that have a poor performance on their own and could only be used in a swarm.

To solve many problems with machine learning algorithms efficiently, it is often necessary to include the changes in the input values over time in the prediction. Furthermore, the domain often requires predictions of targets in the future.  This process is also called time series forecasting. The forecasting timestep $t$ is often called leadtime. Both applications, described in Chapter \ref{energyforecast} and \ref{cyclists}, can be modelled as time series forecasting.

\subsection{Coopetitive Soft Gating Ensemble}
\label{sec:csgemain}
\begin{figure}
    \includegraphics[scale=0.16]{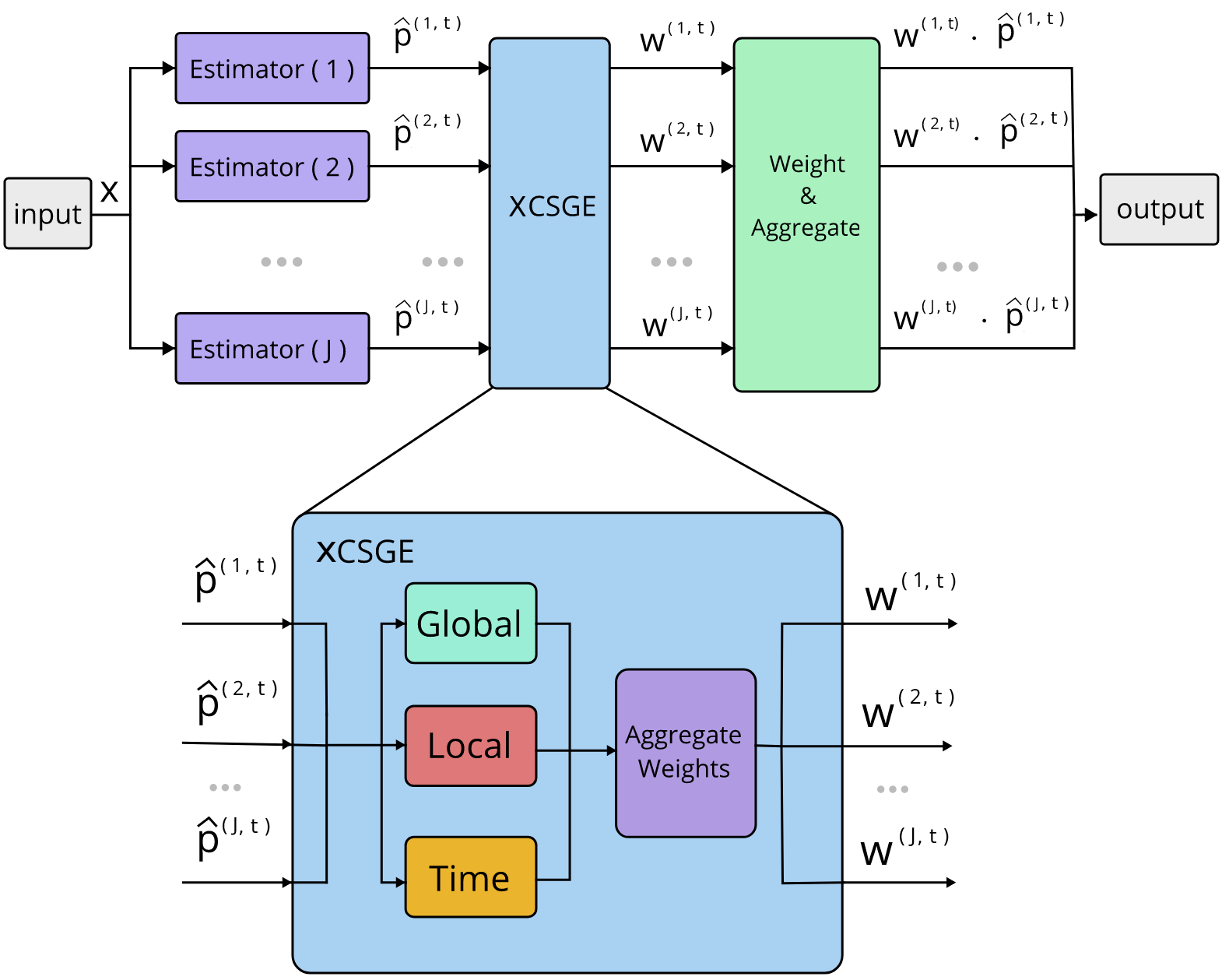}
    \vspace{15px}
     \caption{The architecture of the \textit{XCSGE}. The predictions $\bm{p}^{(j,t)}$ of the input $\bm{x} \in \mathbb{R}^{N_{features}}$ are passed to the \textit{XCSGE} module. The weights are computed regarding \textit{global-}, \textit{local-} and \textit{time dependent weighting}. After that, the predictions are weighted and aggregated for the final prediction.\label{csge_overview_graphic}}
\end{figure}

In this section, the \textit{Extended Coopetitive Soft Gating Ensemble} (\textit{XCSGE}) is introduced. It is a recent ensemble technique which is proposed in \cite{gensler_sick_2017}. In the field of ensemble methods, there are two paradigms to combine individual ensemble members. 
\textit{Weighting} combines all ensemble members in a linear combination, while \textit{gating} selects only one ensemble member. The idea of the \textit{XCSGE} is to gain the possibility to have a mixture of both \textit{weighting} as well as \textit{gating} and let the ensemble itself choose which concept to use for the combination of different predictions. With the \textit{weighting} concept, all ensemble members contribute to the overall prediction - they \textbf{cooperate}. Instead, the \textit{selection} paradigm uses the winner take's it all concept - the ensemble members are in a \textbf{competition}.
 Since the \textit{XCSGE} can work in \textit{selection-} or \textit{weighting} mode, the name \textbf{coopetitive} is a suitcase word which combines the words  \textbf{cooperation} with \textbf{competition}.
Thus, the \textit{XCSGE} is considered to be a \textit{combining classifiers ensemble}, since it uses \textit{strong learners}.
There are no specific requirements on the type of the \textit{ensemble members} learning algorithm, therefore the \textit{XCSGE} is neither a pure \textit{homogenous-}, nor a pure \textit{heterogenous ensemble}.\\
In the following we take advantage of the \textit{Hadamard} operations \cite{hadamard}, which allows us to compute elementwise operations on two vectors.
Fig. \ref{csge_overview_graphic} shows the principle of the \textit{XCSGE} as described in the following. The ensemble includes $J$-ensemble members, with $j \in \{1, ..., J\}$. Each ensemble member  provides multivariate estimations $\bm{p}^{(j,t)} \in \mathbb{R}^M$ for the input $\bm{x} \in \mathbb{R}^{N_{features}}$.
Let $t$ denote the leadtime, with  $t \in \{ 0,...,K\}$, where $K$ is the maximum \textit{leadtime}. Furthermore let $M$ be the number of target variables and $N_{features}$ the number of features.\\
For each prediction the \textit{XCSGE} calculates a \textit{weighting},
regarding three aspects: \textit{global-}, \textit{local-} and \textit{time dependent weighting} and joins them. These three aspects are the core components of the ensemble and are explained in detail later.
After estimating the weighting aspects, all $j$ weights get normalized and each prediction $\bm{p}^{(j,t)}$ gets weighted using the corresponding weight $\bm{w}^{(j,t)}  \in \mathbb{R}^M$.
In the last step, the ensemble's prediction is obtained by aggregation of the weighted predictions. Formally, we can express the weighted aggregation  of the ensembles prediction using the following equation:

\begin{equation}
    \bar{\bm{p}}^{(t)} = \sum^{J}_{j=1} \bm{w}^{(j,t)} \circ \bm{p}^{(j,t)}
\end{equation}

Let $\circ$ denote the \textit{Hadamard product}.
To ensure that the prediction is not distorted we have the following constraint:
\begin{equation}
    \label{formel_norm}
    \sum_{j=1}^{J} w^{(j,t)}_m=1 \qquad \forall t \in \{0,...,K\},\:\forall m \in \{1,...,M\}
\end{equation}

The main goal of the \textit{XCSGE} is to adjust the weights $\bm{w}^{(j,t)}$ optimal regarding the individual performance of the ensemble members.
To compute the strongness, respectively, the weakness of a model, we use error scores, like the \textit{root-mean-squared error (RMSE)}. The performance of an ensemble member is computed by three aspects described in the following.

The \textit{\textbf{global weights}} are calculated for each ensemble member regarding
the overall observed performance of a model during ensemble training. This is a fixed
weighting term. Thereby, overall strong models have more influence than weaker models.
\textit{\textbf{Local weighting}} considers the fact that different ensemble members have diverse prediction quality over the \textit{feature space}. As an example, when considering the problem of renewable energy prediction \cite{gensler_sick_2017}, an ensemble member could perform well on rainy weather inputs but has worse quality when using sunny weather inputs. Therefore, the \textit{local weighting} rewards ensemble members with a higher weighting, that performed well on similar input data.
These weights are adjusted for each prediction during runtime.\\
The \textbf{\textit{time dependent weighting}} aspect is used when performing predictions on timeseries. Sometimes ensemble members perform differently for different leadtimes. When considering the problem of renewable energy prediction again, we can see that the persistence method often achieves superior results in short time horizons, while quickly losing quality for long time predictions. Other methods may perform worse on short time predictions, but have greater stability over time.

To calculate the overall weighting for the $j$-th ensemble member $\bm{w}^{(j,t)}$,
we use the following formular:

\begin{equation}
    \label{overallWeighingFormular}
 \bm{w}^{(j,t)} = \bar{\bm{w}}^{(j,t)} \oslash  \sum_{\tilde{j}=1}^J \bar{\bm{w}}^{(\tilde{j},t)}
 \end{equation}

This equation normalizes the weights and ensures that they sum to one \ref{formel_norm}. Let $\oslash$ denote the \textit{Hadamard division}.
The overall weighting (i.e. including all three weighting aspects) for a specific ensemble member $j$ and target dimension $m$ is given by the following equation:

\begin{equation}
\label{overallweights}
 \bm{\bar{w}}^{(j,t)} = \bm{w}^{(j)}_g \circ \bm{w}^{(j)}_l \circ \bm{w}^{(j,t)}_k
\end{equation}

where $\bm{w}^{(j)}_g \in \mathbb{R}^M$ denotes the \textit{global weighting}, $\bm{w}^{(j)}_l \in \mathbb{R}^M$ the \textit{local weighting} and $\bm{w}^{(j,t)}_k \in \mathbb{R}^M$  the \textit{time dependent weighting}.

\subsection{Soft Gating Principle}
\label{sec:sofgatingformular}
The main goal of the \textit{Coopetitive Soft Gating Ensemble} is to increase the quality of the prediction by weighting solid predictors greater than predictors with fewer quality results for each weighting aspect.
Therefore, a function is implemented which maps the dependency between error of the predictor and its weighting. To do so, the function $\varsigma^{'}_\eta(\Omega, \bm{\rho})$ is used. It is defined as follows:

\begin{equation}
\varsigma^{'}_\eta(\Omega, \bm{\rho}) = \sum_{j=1}^{J} \bm{\Omega_j} \oslash ( \bm{\rho}^{\circ \, \eta} + \begin{pmatrix} \hat{\epsilon}_1 \\ \vdots \\ \hat{\epsilon}_M \end{pmatrix}) , \eta \in \mathbb{R}^{+}
\end{equation}

Let $\Omega$ be a set which contains the reference errors of all $J$ ensemble members, and $\bm{\rho} \in \mathbb{R}^{M}$ be the error of the ensemble member that is to be weighted.\\
Let $\circ$ denote the \textit{Hadamard exponentiation}. $\hat{\epsilon}$ is a small constant to prevent a division by zero.
The parameter $\eta$ is chosen by the user. It controls the linearity of the weighting. For greater $\eta$ the \textit{XCSGE} tends to work as a \textit{selecting ensemble}, whereas smaller $\eta$ result in a \textit{weighting ensemble}.

\begin{figure}
  \centering
    \includegraphics[scale=0.55]{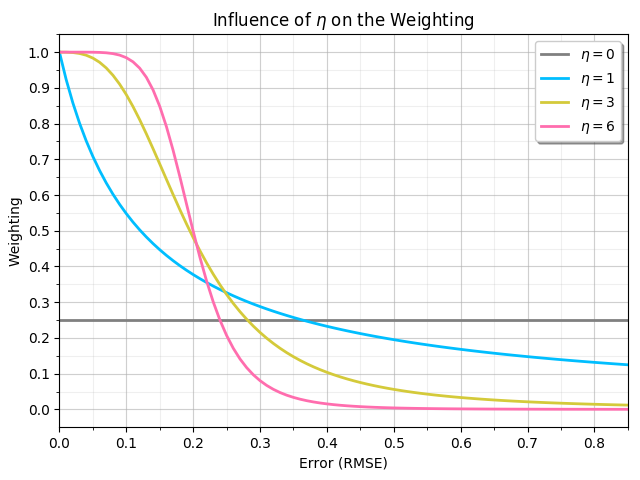}
      \vspace{0px}
      \caption{\label{etaRmseGraphic} The error (RMSE) of a predictor is drawn on the x-axis, while the y-axis contains the corresponding weights computed by $\varsigma^{'}_\eta$. For greater $\eta$ a higher error gets more regulated with less weighting than for smaller $\eta$. }
\end{figure}

By taking a closer look on $\varsigma^{'}_\eta$, we can discover the following characteristics:

\begin{itemize}
\item The function $\varsigma^{'}_\eta$ is falling monotonously (see Fig. \ref{etaRmseGraphic}). 
\item For greater $\bm{\rho}$ or the error of an ensemble member, $\varsigma^{'}_\eta$ returns smaller weightings.
\item For $\eta = 0$ every ensemble member is weighted with $\frac{1}{J}$, disrespecting the error of it's prediction.
\end{itemize}

To ensure that  $\sum_{j=1}^{J} w^{(j)}_m = 1 \; \quad \forall m \in \{1, ..., M\}$, $\;\varsigma^{'}_\eta(\Omega, \rho)$ is normalized in the following way:

\begin{equation}
\label{softGatingFormel}
\varsigma^{}_\eta(\Omega, \bm{\rho}) = \varsigma^{'}_\eta(\Omega, \bm{\rho}) \oslash \sum_{j=1}^{J}\varsigma^{'}_\eta(\Omega, \bm{\Omega_{j}})
\end{equation}

Besides the fact of having only one parameter ($\eta$), the soft gating function offers the advantage of operating directly on errors of the ensemble members and, therefore, directly links to the actual data.

\vspace*{10px}

\subsection{Global Weighting}
\label{sec:global}
The \textit{global weighting} is calculated during ensemble training and then remains constant. Ensemble members that performed well on the training data get greater weights compared to those that performed worse.
Therefore, the difference between estimation and ground truth is calculated with

\begin{equation}
\label{formelDiff}
\bm{e}_n^{(j)} = \sum_{\bar{t}=0}^{K}S(\bm{p}_n^{(j,\bar{t})},\bm{y}_n) \qquad \text{with } 1 \leq n < N_{samples}
\end{equation}

Let $\bm{p}_n^{(j,t)} \in \mathbb{R}^M$ be the prediction of the $j$-th ensemble member on leadtime $t$ and $\bm{y}_n \in \mathbb{R}^M$ be the corresponding ground-truth.\\
Let $S$ be an arbitrary error metric, which for example could be the \textit{root-mean-squared-error (RMSE)} for regression or the \textit{Log Loss} for classification.
The only condition that has to hold is that it has to be falling monotonously with increasing errors to work properly with the \textit{soft gating principle}, see Eq. \ref{softGatingFormel}.
Since the \textit{global weighting} is independent of leadtime, we sum over all leadtimes.
The error score $\bm{R}^{(j)} \in \mathbb{R}^M$ of the $j$-th ensemble member is calculated by:

\begin{equation}\label{formelGlobaleGewichtungR}
 \bm{R}^{(j)} = \frac{1}{N_{samples}} \cdot \sum_{n=1}^{N_{samples}} \bm{e}_n^{(j)}
\end{equation}

\begin{equation}
 \bm{R} = \{ \bm{R}^{(1)},... , \bm{R}^{(j)},... , \bm{R}^{(J)} \}
\end{equation}

The set $\bm{R} \in \mathbb{R}^M$ contains all error scores of the $J$ ensemble members. By that, the \textit{global weight} of the $j$-th ensemble member is calculated by

\begin{equation}
\bm{w_g}^{(j)} = \varsigma^{}_{\eta_{global}}(\bm{R}, \bm{R}^{(j)})
\end{equation}

We use the parameter $\eta_{global}$ for the soft gating principle. $\eta_{global}$ is chosen during ensemble training, as described in Section \ref{sec:training}.

\subsection{Local Weighting}
\label{sec:local}
The \textit{local weighting} considers the quality difference between the predictors for distinct situations over the whole \textit{feature space}. Therefore, the \textit{local weighting} rewards \textit{ensemble members} with a higher weighting, that performed well on similar input data.
In contrast to the \textit{global weighting}, the \textit{local weighting} is calculated for each ensemble member and each prediction at runtime. First, we use the training set $X_H$ and make predictions for all ensemble members. Afterward, we calculate the error per sample using the scoring function $P$.
Then, we train an arbitrary regression model $M_{local}^{(j)}$ per ensemble member, which has the \textit{feature space} as input and the previously calculated error as output.
The regressions model $M_{local}^{(j)}$ estimates the expected error for the respective \textit{ensemble member} based on the input. As a intuitive approach, we use a \textit{k-nearest neighbor regressor} \cite{knn_reg}, unless otherwise stated.
To calculate, the local error score $\bm{q}^{(j)} \in \mathbb{R}^M$ for an ensemble member $j$ is given by 

\begin{equation}\label{knn_variante}
 \bm{q}^{(j)} = M_{local}^{(j)}(\bm{x})
\end{equation}

In the following, we calculate the weighting analogous to the \textit{global weighting} by using the soft gating formula $\varsigma_{\eta}$

\begin{equation}
\bm{Q} = \{ \bm{q}^{(1)}, ... , \bm{q}^{(j)} , ... , \bm{q}^{(J)}\}
\end{equation}

$\bm{Q} \in \mathbb{R}^M$ contains all local error scores of the $J$ ensemble members.
Then, the \textit{local weight} $\bm{w}_l^{(j)}$ is calculated by using the soft gating principle:

\begin{equation}
\bm{w}_l^{(j)} =  \varsigma^{}_{\eta_{local}}( \bm{Q}, \bm{q}^{(j)})
\end{equation}

\subsection{Time Dependent Weighting}
\label{sec:time}
The \textit{time dependent weighting} considers the fact that the quality of an ensemble member may vary for different leadtimes.
$\hat{\bm{P}}^{(j, K)}_n \in \mathbb{R}^M$ contains all predictions of training sample $n$ of ensemble member $j$ for a specific leadtime $t \in \{0,...,K\}$.

\begin{equation}
\label{timeseriesSet}
\hat{\bm{P}}^{(j, K)}_{n} = \{ \bm{p}_{n}^{(j,0)},\bm{p}_{n}^{(j,1)}
 ... , \bm{p}_{n}^{(j,K)} \}
\end{equation}

The error for a specific leadtime $t$ is calculated by averaging the error over all training samples:

\begin{equation}
\bm{R}^{(j,t)} = \frac{1}{N_{samples}} \cdot \sum_{n=1}^{N_{samples}} S(\bm{p}_n^{(j,t)},y_n)
\end{equation}

With $\bm{p}_n^{(j,t)} \in \hat{\bm{P}}^{(j, K)}$ and $y_n$ being the corresponding ground truth. It holds $\bm{R}^{(j,t)} \in \mathbb{R}^M$. To calculate the error score for leadtime $t$ of ensemble member $j$, we use the following equation:
 
\begin{equation}
\bm{r}^{(j,t)} = \frac{\bm{R}^{(j,t)}}{ \sum_{\tilde{t} = 0}^{T} \bm{R}^{(j,\tilde{t})}}
\end{equation}

$\bm{r}^{(j,t)} \in \mathbb{R}^M$ is a score that compares the error of the prediction with the leadtime $t$ to the average error in the leadtime interval $\{0,...,t, ... ,K\}$. The weight $\bm{w}_k^{(j,t)} \in \mathbb{R}^M $ is calculated analogous to \textit{global-} and \textit{local weighting} using the soft gating principle

\begin{equation}
\bm{P}^{(t)} = \{ \bm{r}^{(1,t)} ,..., \bm{r}^{(j,t)}, ..., \bm{r}^{(J,t)} \} 
\end{equation}

\begin{equation}
\bm{w}_k^{(j,t)} = \varsigma^{}_{\eta_{time}}(\bm{P}^{(t)}, \bm{r}^{(j,t)})
\end{equation}

We use the parameter $\eta_{time}$ for the soft gating principle. $\eta_{time}$ is chosen during ensemble training as described in Section \ref{sec:training}.

\subsection{Model Fusion and Ensemble Training}
\label{sec:training}
To calculate the output of the $J$ ensemble members, we use the following equation:

\begin{equation}
    \bar{\bm{p}}^{(t)} = \sum^{J}_{j=1} \bm{w}^{(j,t)} \circ \bm{p}^{(j,t)}
\end{equation}

As mentioned in Section \ref{sec:sofgatingformular}, the parameter $\eta$ is chosen by the user and controls the non-linearity of the system. Since there are three aspects, \textit{global-, local-} and \textit{time dependent weighting}, it follows that there are also three $\bm{\eta} = (\eta_{global}, \eta_{local}, \eta_{time})$ to be chosen. Let $f_{XCSGE}(\bm{x}_n,\bm{\eta)}$ be the prediction of the \textit{XCSGE} for the input sample $x_n$ with the set of $\bm{\eta}$.
Then the following minimization problem solves the task to adjust the set of $\bm{\eta}$:
\begin{equation}
\label{FormelCSGEMin}
\sum^{N_{samples}}_{n=1}[\bm{y}_n - f_{XCSGE}(\bm{x}_n,\bm{\eta})]^2 + c \cdot \sum^{3}_{s=1}\eta_s
\end{equation}

Where $\sum^{N_{samples}}_{n=1}[\bm{y}_n - f_{XCSGE}(\bm{x}_n,\bm{\eta)}]^2$ are the summed errors over the training data, while $c \cdot \sum^{3}_{s=1}\eta_s$ is a regularisation term to control overfitting.

\subsection{Time Lagged Ensemble}
\label{sec:time_lagged}

In the following, we will discuss a technology that is particularly interesting for time-critical applications, especially when several sensors measure data. In practice there is often a delay in the arrival of the data in the prediction models. 
A possible approach is to train the models on time-delayed training data. 
In this approach, the ground-truth is shifted forward by one time unit at a time. All in all, this results in $t \cdot j$ ensemble members, where $j$ is the number of different ML models, and $t$ is the number of time units by which a shift should take place.\\
Since the weighting is recalculated and normalized for each prediction, it is also possible to exclude specific time-shifted models from the ensemble prediction. This is of interest if a time delay of a sensor is known to compensate for this delay.

\begin{figure}[H]
  \centering
    \includegraphics[scale=0.45]{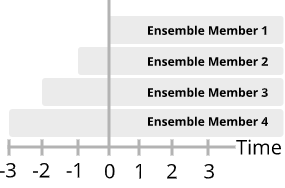}
      \vspace{10px}
      \caption{\label{timeLagged} \small Four ensemble members, each shifted by one time unit.}
\end{figure}

\section{Renewable Energyforecast}
\label{energyforecast}

In the following, we investigate the \textit{XCSGE} on its application on renewable power forecast.
Power forecasts on wind farms using the \textit{CSGE} has already been investigated in \cite{gensler_sick_probabilistic}, \cite{gensler_knn_approx} and \cite{gensler_sick_2017}.
Power forecasts are an essential tool to schedule the power supply.
Wind farms, as well as solar farms power generation demands on volatile weather situations, which makes them hard to predict.
Since the proportion of renewable energy is growing strongly, better prediction models are necessary to ensure power grid stability.
In this section, the \textit{XCSGE} is used to predict the power generation of wind farms (see Section \ref{windfarms}) as well as solar farms (see Section \ref{solarfarms}).
We use the meteorological properties of weather forecasts as features to build models on them.
The problem can be modeled as a regression task. 
On the one hand, the difficulty of the problem lies in the non-linear relationship between weather and power generation.
On the other hand, the quality of the weather forecasts, and thus the quality of the input features decreases over leadtime.
Since the \textit{XCSGE} considers the quality fluctuation of the ensemble members over leadtime, it is suitable for this task.\\
Next to an evaluation on the datasets, a comparison with other state-of-the-art ensemble methods, such as the \textit{stacking} ensemble method, is given.
For both forecasting problems, the same type of ensemble members are trained, as detailed in Section \ref{sec:csge_renewable_base learners}.
Next to the ensemble members, the training process of the \textit{XCSGE} itself is identically for both wind farms and solar farms. It can be found in Section \ref{sec:csge_renewable_training}.
The description of the datasets as well as the preprocessing and evaluation is given in the corresponding sections, since they are diverse for each forecasting problem.

\subsection{Ensemble Members}
\label{sec:csge_renewable_base learners}
For each wind farm respectively solar farm, four ensemble members are trained using the \textit{structure diversity} principe. The trained modells are: \textit{support vector regression (SVR)} \cite{Scholkopf:2001:LKS:559923}, \textit{linear regression (ridge regression)} \cite{doi:10.1080/00401706.1970.10488634}, \textit{neuronal network (MLP)} \cite{159058} as well as a \textit{gradient boost regressor tree (GBRT)} \cite{friedman2001} .
For the sake of runtime, the ensemble members are trained with fixed parameters.

\subsection{Ensemble Training}
\label{sec:csge_renewable_training}

The ensemble members as well as the \textit{XCSGE} and \textit{stacking} are evaluated by using a ten-fold cross-validation. In each fold, the data is split in a training set (90\%) and a test set (10\%).
The test set is not shuffled to keep the structure of time indices.
Furthermore, the training fold is split into a base learner set (70\%) and into an ensemble set (30\%).
The base learner set is used to train the four ensemble members, while the ensemble set is used to train the \textit{XCSGE} and \textit{stacking}.
For \textit{stacking} a \textit{linear regression} as well as an \textit{MLP} as \textit{metalearner} is evaluated.
A gridsearch optimizes the hyperparameters of the ensemble methods.
We use a \textit{k-nearest neighbor regressor} as the machine learning model $M_{local}^{(j)}$ to calculate the \textit{local weighting}.
For the \textit{XCSGE} a gridsearch on the parameter $k$ is executed, which controls the number of nearest neighbors for the \textit{k-nearest neighbor regressor} $M_{local}^{(j)}$.
The parameter $k$ is chosen in the range $\{9, 50, 100\}$.
The set of $\eta = (\eta_1, \eta_2, \eta_3)$ is optimized as pointed out in the training process of the \textit{XCSGE}.
For the \textit{stacking} with the \textit{MLP} as metalearner the possible hidden layer sizes are optimized using a gridsearch. The hidden layer sizes are chosen in a range of $[50 - 100]$.
For the \textit{linear regression} \textit{metalearner} there is no need to optimize the model parameters.
The \textit{mean squared error} is used as the \textit{XCSGE}'s errorfunction  $S$.

\subsection{Wind farms}
\label{windfarms}

In this chapter, the \textit{XCSGE} is applied to wind farm datasets.
First, a short overview of the structure of the data, as well as general information, is given in Section \ref{sec:method_windfarm}. After that, Section \ref{sec:pre_windfarm} gives a summary of the preprocessing.
The training process of the machine learning models, which are used as the ensemble members of the \textit{XCSGE} as well as the \textit{stacking}, is pointed out in Section \ref{sec:csge_renewable_base learners}.
The evaluation method, as well as the ensemble training, can be found in Section \ref{sec:csge_renewable_training}.
A presentation of the final results of the experiments is given in Section \ref{wind:evaluation} and \ref{wind:evaluation_local}.

\subsubsection{Dataset}
\label{sec:method_windfarm}

The \textit{XCSGE} is evaluated on 70 different wind farms.
The wind farms are spread all over Europe, containing both onshore and offshore wind farms.
Every wind farm dataset contains measured power generation, which is captured every hour on two consecutive years resulting in a maximum of 17520 measured samples. Nevertheless, some wind farms only have $50\%$ data samples of the two years captured.
The measured generated power of the wind farm is averaged hourly by the maximum power of the wind farm.
Besides the measured generated power of the wind farm, the datasets contain a corresponding day-ahead weather forecast for the wind farm's location.
The day-ahead weather forecast is updated every day and it covers the weather situation in one hour steps up to 24 hours for the next day.
The weather forecast contains seven methereological features which are:

\begin{itemize}
\item Air Pressure
\item Humidity
\item Temperature
\item Wind Direction \{Zonal, Meridional\} 100m
\item Wind Speed \{10m, 100m\}
\end{itemize}

\subsubsection{Preprocessing}
\label{sec:pre_windfarm}
First, every feature is standardized \cite{statisticbook} while the ground-truth is normalized by the maximum power.
The scaling of the ground-truth (power generation) is necessary to compare the error of the wind farms among themselves in Section \ref{wind:evaluation}.
Furthermore, the four features \textit{wind speed 100m}, \textit{wind speed 10m}, \textit{wind directional zonal 100m} and \textit{wind directional merdidonal 100m} are time shifted up to two time units both in the future as well as in the past (+/-2 hours) to model the time dependency between the features.
After the preprocessing, the datasets contain 30 features.
Since the weather model gets updated every 24 hours, the \textit{timedependent weightings} property $k$ is $(k_{min}=25, k_{max}=48, \Delta=1)$.\\

\subsubsection{Analysis of local weighting models}
\label{wind:evaluation_local}

\begin{figure}[h!]
  \centering
  
    \includegraphics[scale=0.5]{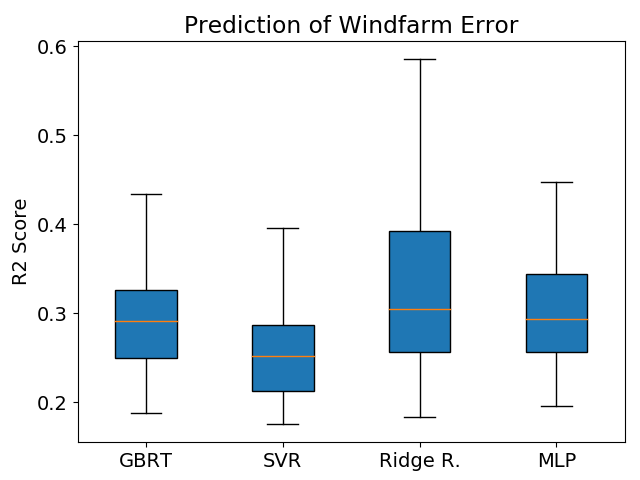}
\vspace{10px}
     \caption{\footnotesize R2-score of the \textit{local weighting} models $M_{local}^{(j)}$.}
    \label{error_local_wind}
\end{figure}

To calculate the \textit{local weighting}, we take advantage of models that estimate the expected error of an ensemble member for a specific sample.
For this purpose we use a \textit{k-nearest neighbor regressor} as the local error model $M_{local}^{(j)}$.
We evaluated these models within a six-fold cross-validation. The results can be seen in Figure \ref{error_local_wind}. The local error models are equally good at predicting the error for their respective ensemble member with an R2-score of about 0.3

\subsubsection{Evaluation}
\label{wind:evaluation}
The models are evaluated on the test set using a ten-fold cross-validation.
The setup of the folds is pointed out in Section \ref{sec:csge_renewable_training}. Tables \ref{table_windfarm_mse} and \ref{table_windfarm_r2} show the \textit{RMSE} as well as the corresponding \textit{R2-score} of all 70 wind farms averaged over all ten folds.
Next to the mean, the skill score is calculated, which is a percentage improvement compared to a reference model. In this case, the worst model (\textit{ridge regression}) is chosen to be the reference model. The results are visualized by a boxplot in Figure \ref{box_wind}.
On both error metrics, the \textit{XCSGE} performs the best achieving a skill score of 11.01\% on the \textit{RMSE} and 10.24\% on the \textit{R2-score}.

\begin{figure}[h!]
  \centering
  
    \includegraphics[scale=0.23]{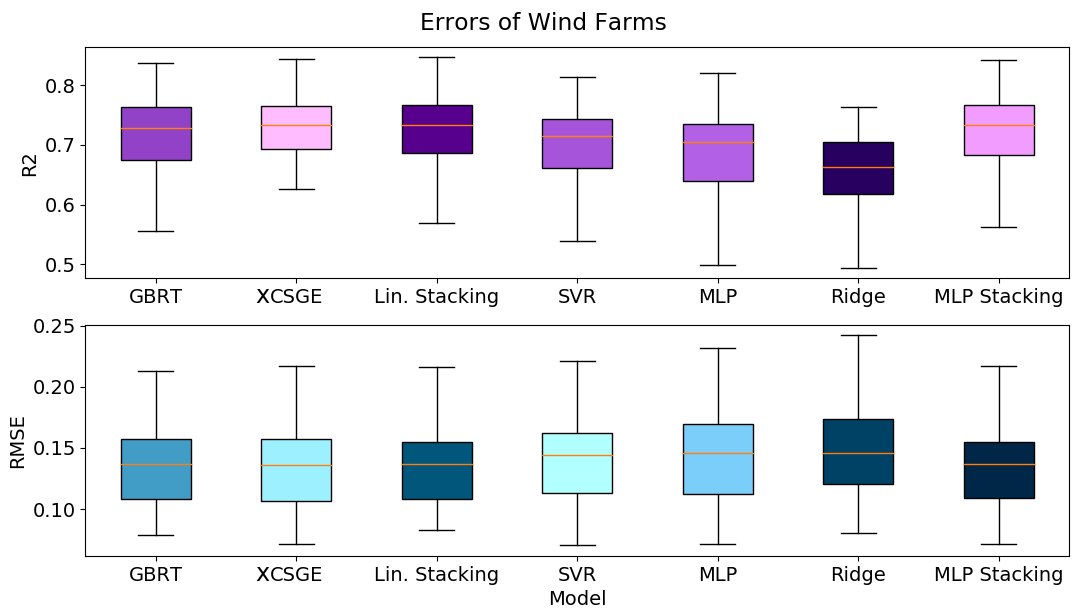}
    \vspace{0px}
     \caption{\footnotesize Box plot of \textit{RMSE} and \textit{R2-score} of wind farms.}
    \label{box_wind}
\end{figure}

\resizebox{!}{.032\paperheight}{
\begin{tabular}{lcccc|ccc}
\toprule
 &   GBRT &    MLP &    SVR &  Ridge R. &   XCSGE &  MLP Stacking &  Linear Stacking \\
\midrule
Mean & 0.1365 & 0.1407 & 0.1439 & 0.1516 & \textbf{0.1349} & 0.1356 & 0.1356\\
Variance &    0.0013 &    \textbf{0.0011} &    0.0014 &    0.0017 &    0.0012 &    0.0012 &    0.0012\\
Minimum &    0.0783 &    \textbf{0.0704} &    0.07107 &    0.0799 &    0.0711 &    0.0711 &    0.0827\\
Maximum    & 0.2664 &    0.2606 &    0.2777 &    0.3015 &    0.2608 &    0.2610 &    \textbf{0.2587}\\
Skill Score & 9.92\% & 7.18\% & 5.08\% & 0.0 \% & \textbf{11.01\%} & 10.54\% & 10.55\% \\ 

\bottomrule
\end{tabular}
}
\vspace{10px}
\captionof{table}{\footnotesize \textit{RMSE} of 10-fold cross-validation. The best score for each wind farm is highlighted bold.} 
\label{table_windfarm_mse}
\vspace{10px}

\resizebox{!}{.03\paperheight}{
\begin{tabular}{lcccc|ccc}
\toprule
 &   GBRT &    MLP &    SVR &  Ridge R. &   XCSGE &  MLP Stacking &  Linear Stacking \\
\midrule
Mean & 0.6614    & 0.6430 &    0.6309 &    0.6095 &    \textbf{0.6719}    & 0.6652 &    0.6608 \\
Variance &    0.0861    & 0.0774    &0.0763    &\textbf{0.0342}    &0.0726    &0.0856    &0.1013\\
Minimum &    -1.2140&    -1.0909&    -1.0162&    \textbf{-0.2478}&    -0.9674&    -1.1180&    -1.4362\\
Maximum    & 0.8370    &0.8144    &0.8198    &0.7638    &0.8442    &0.8417    &\textbf{0.8465}\\
Skill Score & 8.52\% & 5.489\% & 3.502\% & 0.0\% & \textbf{10.237\%} & 9.13\% & 10.08\% \\ 
\bottomrule
\end{tabular}
}
\vspace{10px}
\captionof{table}{\footnotesize R2-score of 10-fold cross-validation. The best score for each wind farm is highlighted bold.} 
\label{table_windfarm_r2}

\vspace{10px}

\begin{figure}[h!]
  \centering
  
    \includegraphics[scale=0.26]{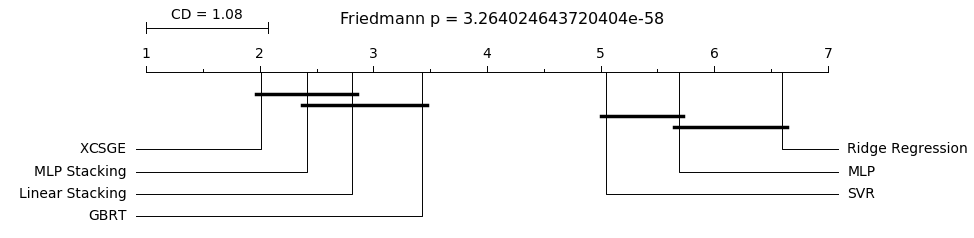}
\vspace{5px}
     \caption{\footnotesize Evaluation of ranked performance of the models using the \textit{Nemenyi post-hoc test}. As can be seen from the figure, the \textit{XCSGE} is significantly better than its ensemble members.}
    \label{friedmann_wind}
\end{figure}

The ranked performance of the algorithms among all of the 70 wind farms is furthermore analyzed using the \textit{Friedman test} \cite{friedmann} in conjunction with the \textit{Nemenyi post-hoc test} \cite{friedmann}. The results are given in Fig. \ref{friedmann_wind}. The \textit{Friedman p} value given in
the figure indicates that the ranks are significantly different, using a significance level of $\alpha=0.05$. As it can be seen from the \textit{Nemenyi test}, the \textit{stacking ensembles} have not a significantly better ranked performance in comparison
to the ensemble members. However, the \textit{XCSGE} achieved a significantly better performance than the ensemble members, which is not given for the other ensemble methods. 

\subsection{Solar farms}
\label{solarfarms}

In this section, the \textit{XCSGE} is applied to solar farms to predict the power forecast.
First, we will give a short overview of the data in Section \ref{solar:dataset}, followed by a summary of the preprocessing in Section \ref{solar:preprocessing}.
The training process of the machine learning models, which are used as the ensemble members of the \textit{XCSGE} as well as the \textit{stacking} is pointed out in Section \ref{sec:csge_renewable_base learners}
The evaluation method, as well as the ensemble training, can be found in Section \ref{sec:csge_renewable_training}.
Finally, the section concludes with a short evaluation of the trained models in Section \ref{solar:evaluation} and \ref{solar:evaluation_local}.\\

\subsubsection{Dataset}
\label{solar:dataset}

We also evaluated the \textit{XCSGE} on 114 different solar farms, which are spread all over Europe.
In comparison to wind farms, every sample is captured in a three hour circle. Every dataset contains measured power generation, which is captured in three hour circles on two consecutive years resulting in 3871 measured samples.
The measured generated power of the solar farm is averaged over three hours.
Next to the generated power, the dataset also contains a corresponding day-ahead weather forecast for the solar farm's location.\\ The day-ahead weather forecast is updated on a daily basis, and it covers the weather situation in one hour steps up to 24 hours for the next day.\\
The weather forecast contains 50 meteorological features, including:

\begin{itemize}
\item Sun Position \{Theta Z, Extra Terr, Solar Height\}
\item Clear Sky \{Direct, Global\}
\item Relative Humidity At 0
\item Net Solar Radiation \{Direct, Diffuse\}
\item Snow \{depth, fall\}
\end{itemize}

\subsubsection{Preprocessing}
\label{solar:preprocessing}

First, every feature is standardized while the ground-truth is normalized by the maximum power.
The scaling of the ground-truth (power generation) is necessary to compare the error of the solar farms among themselves in Section \ref{solar:evaluation}.\\
To model the time dependency between the features, the features are time shifted .
The time shift is done both in the past as well as in the future up to one time unit (+/- 3 hours)\\
After the preprocessing, our dataset contains around 65 features.
The weather model is updated every 24 hours leading to the \textit{timedependent weightings} property $(k_{min}=25, k_{max}=48, \Delta=3)$.

\vspace{10px}
\subsubsection{Analysis of local weighting models}
\label{solar:evaluation_local}

\begin{figure}[h!]
  \centering
  
    \includegraphics[scale=0.45]{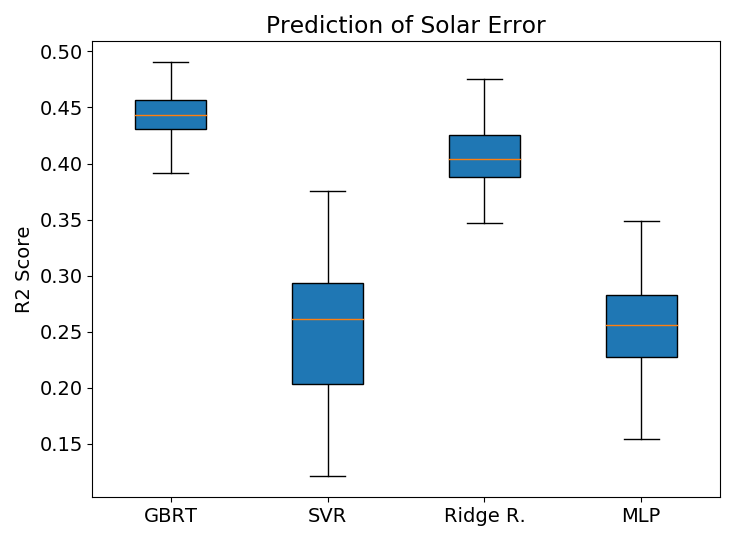}
\vspace{10px}
     \caption{\footnotesize R2-score of the \textit{local weighting} models $M_{local}^{(j)}$.}
    \label{error_local_solar}
\end{figure}

For the calculation of the \textit{local weighting}, we also use a \textit{k-nearest neighbor regressor} as the local error model $M_{local}^{(j)}$.
We evaluated these models within a six-fold cross-validation. The results can be seen in Figure \ref{error_local_solar}. The local error models for the \textit{GBRT} and \textit{ridge regression}  
respectively achieve significantly better results in predicting the error compared to the two models for \textit{SVR} and \textit{MLP}.

\subsubsection{Evaluation}
\label{solar:evaluation}
\begin{figure}[h!]
  \centering
  
    \includegraphics[scale=0.22]{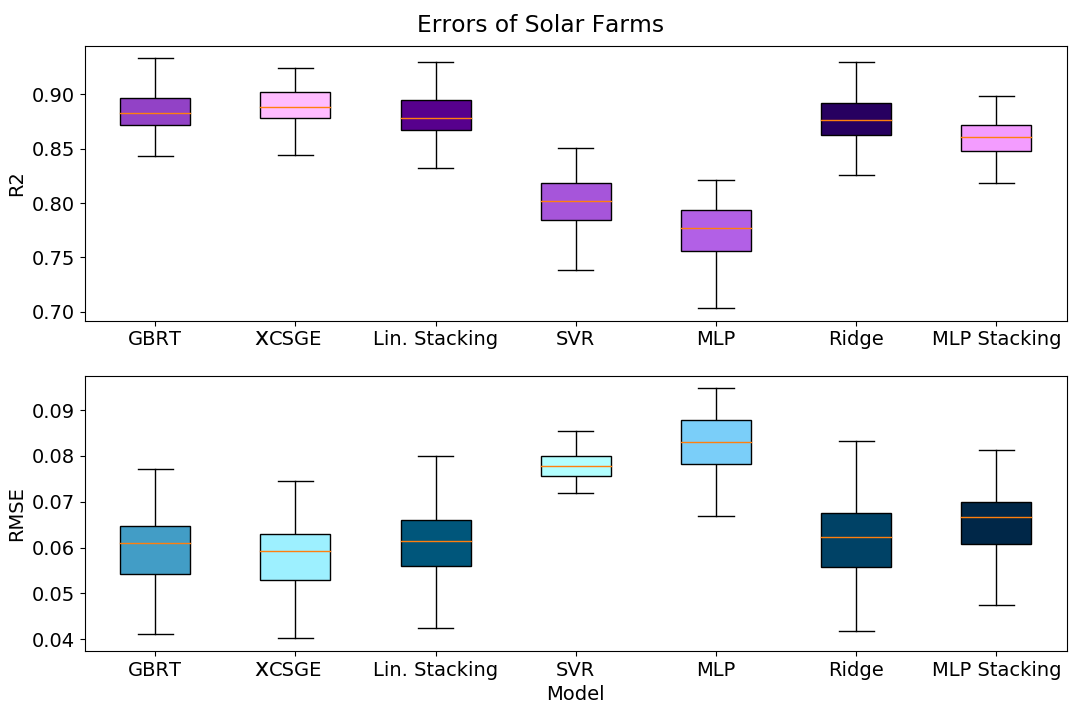}
\vspace{5px}
     \caption{\footnotesize Box plot of \textit{RMSE} and \textit{R2-score} of solar farms.}
    \label{box_solar}
\end{figure}

Table \ref{table_solarfarm_mse} and \ref{table_solarfarm_r2} show the \textit{RMSE} and the \textit{R2-score} of all 114 solar farms averaged over all ten folds. The setup of the folds is pointed out in Section \ref{sec:csge_renewable_training}. The full tables can be found in the Appendix \ref{appendix_wind_1}, \ref{appendix_wind_2}, \ref{appendix_solar_1} and \ref{appendix_solar_2}. Next to the mean, we also calculated the skill score using the worst model as reference. We can see, that the \textit{XCSGE} performs the best achieving a skill score of 30.00\% on \textit{RMSE} and 15.27\% on the \textit{R2-score}.
In addition, the results are visualized by a boxplot in Figure \ref{box_solar}.
When considering the ensemble members \textit{RMSE} and \textit{R2-score} we can observe, that \textit{GBRT} as well as the \textit{ridge regression} performs the best over nearly all solar farms. The \textit{MLP} and the \textit{SVR} have the worst scores on all wind farms.\\
When comparing the \textit{R2-score} of the \textit{ridge regression} on the wind farm datasets (0.60955) with the one on the solar farm datasets (0.8751), we can assume that the correlation between weather and energy yield is more linear for solar farms compared to wind farms.

\resizebox{!}{.03\paperheight}{

\begin{tabular}{lcccc|ccc}
\toprule
&  GBRT &  MLP &   SVR &  Ridge R. &  XCSGE &  ANN Stacking &  Linear Stacking \\
\midrule
Mean & 0.0595    & 0.0779 &    0.08280 &    0.0611 &    \textbf{0.0579}    & 0.0651 &    0.0606 \\
Variance &    $6.53E^{-5}$    & \textbf{$\boldsymbol{1.33E^{-5}}$}    & $4.56E^{-5}$    & $8.08E^{-5}$    & $6.051E^{-5}$    &$5.961E^{-5}$    &$6.69E^{-5}$\\
Minimum &    0.0411&    0.0690&    0.0610&    0.0417&    \textbf{0.0401}& 0.0475 & 0.0423\\
Maximum    & 0.0909    &0.0975    &0.1067    &0.0921    &\textbf{0.0869}    &0.0946    & 0.0912\\
Skill Score & 28.08\% & 5.86\% & 0.00 \% & 26.12 \% & \textbf{30.00\%} & 21.33\% & 26.71 \% \\ 
\bottomrule
\end{tabular}
}
\vspace{10px}
\captionof{table}{\footnotesize\textit{RMSE} of 10-fold cross-validation for each solar farm.The best \textit{RMSE} score for each solar farm is highlighted bold.} 
\label{table_solarfarm_mse}

\vspace{5px}
\resizebox{!}{.032\paperheight}{

\begin{tabular}{lcccc|ccc}
\toprule
 &  GBRT &  MLP &   SVR &  Ridge R. &  XCSGE &  ANN Stacking &  Linear Stacking \\
\midrule
Mean & 0.8819    & 0.7950 &    0.7704 &    0.8751 &    \textbf{0.8881}    & 0.8587 &    0.8775 \\
Variance &    0.0005    & 0.0013    & 0.0013    & 0.0009    & \textbf{0.0004}    & 0.0005    & 0.0005\\
Minimum &    0.7804 &    0.6865 &    0.6102 &    0.6886 &    \textbf{0.7929} & 0.7559 & 0.7675\\
Maximum    & 0.9332    &0.9002    &0.8795    &0.9298    &\textbf{0.9392}    &92412    &0.9301\\
Skill Score & 14.47\% & 3.18\% & 0.00 \% & 13.58 \% & \textbf{15.27\%} & 11.46\% & 13.89 \% \\ 
\bottomrule
\end{tabular}
}
\vspace{10px}
\captionof{table}{\footnotesize\textit{R2-score} of 10-fold cross-validation for each solar farm. The best \textit{RMSE} score for each solar farm is highlighted bold.} 
\label{table_solarfarm_r2}

\vspace{10px}

\begin{figure}[h!]
  \centering
  
    \includegraphics[scale=0.26]{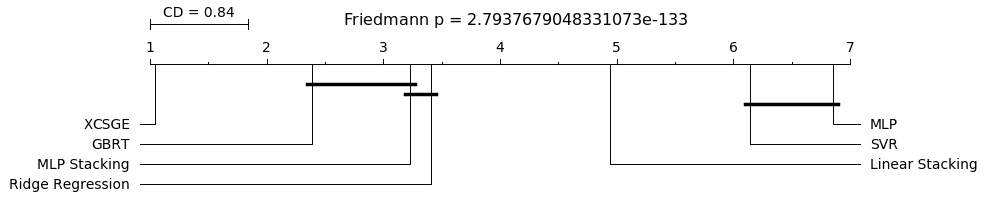}
\vspace{5px}
     \caption{\footnotesize Evaluation of the ranked performance of the models using the \textit{Nemenyi post-hoc test}. As can be seen from the figure, the \textit{XCSGE} has the best ranked performance and is significantly better than the other comparison models.}
    \label{friedmann_solar}
\end{figure}

The ranked performance of the algorithms among all of the 114 solar farms is furthermore analyzed using the \textit{Friedman test} in conjunction with the \textit{Nemenyi post-hoc test}. The results are given in Fig. \ref{friedmann_solar}. The \textit{Friedman p} value, given in
the figure, indicates that the ranks are significantly different using a significance level of $\alpha=0.05$. As can be seen from the \textit{Nemenyi test}, the \textit{XCSGE} has a significantly better ranked performance in comparison
to the other models.

\section{Cyclists Basic Movement Detection}
\label{cyclists}
In urban traffic situations, there are many different road users e.g. cars, buses, cyclists. Especially cyclists are very vulnerable since they are easily overseen by other road users.
Therefore, \textit{advanced driver assistance systems} can help the driver of a vehicle to anticipate possible dangerous situations and execute a stop to avoid a potential collision \cite{PDAD1}.
To do so, those systems need to detect the current movement state, e.g. waiting and moving, and subsequently need to forecast the cyclist's future trajectory.
To gain information about the cyclist's movement, we use the sensors of smart devices.
Smart devices such as smartphones or smartwatches have seen an enormous increase in performance over the last few years. The increase in performance is not only limited to battery life and computing power but also the precision of the contained sensors is continuously increasing. Therefore, they are an ideal way to gain information as nearly everybody has one with them all the time. It is of particular challenge to develop models from this information that can be used to detect the current cyclist movement state. In the following, we refer to the movement state as movement primitives.
Data from multiple smart devices, e.g. smartphone, smartwatch, and a smart sensor-equipped helmet, can be combined to quickly detect the cyclist current movement state and communicate it to nearby intelligent vehicles using G5 \cite{AAET}. This is especially helpful to resolve occlusion situations, e.g. cyclist is entering the road occluded by another vehicle or object.\\
In this section, we will use the \textit{XCSGE} to classify the movement primitives of cyclists.
The difficulty of the problem results from different aspects.
On the one hand the dimensionality of the feature space is extremely high and for the considered amount of data is extremely large (738 features and 210452 samples).
There are several sensors used, which capture a great number of different physical properties e.g. acceleration.
On the other hand, we have to track a lot of samples over a relatively short time horizon for a precise classification of the cyclist's movement primitive.
The great amount of features and data requires an appropriate preprocessing to develop strong models.

\subsection{Dataset}
\label{cyclist_method}
\begin{figure}[h!]
  \centering
  
    \includegraphics[scale=0.35]{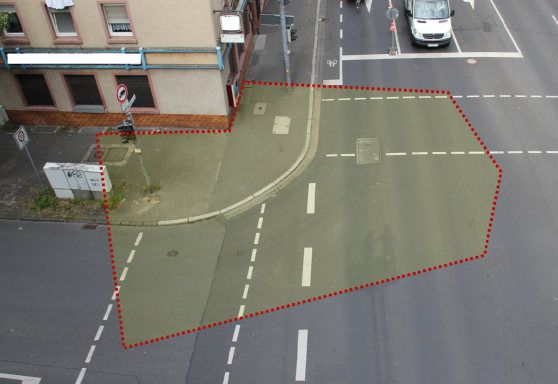}
\vspace{10px}
     \caption{\footnotesize Joint viewport of the two cameras.}
    \label{crossing_camera}
\end{figure}

The considered dataset is captured in several experiments on a public crossing in Aschaffenburg. A picture of the crossing can be seen in Fig. \ref{crossing_camera}.
In total, 50 different cyclists took part in the experiments.
Every cyclist was equipped with a Samsung Galaxy S6 smartphone and two Motorola Moto 360 smartwatches, which tracked the data:

\begin{itemize}
\setlength\itemsep{0.5em}
\item \textbf{Smartphone} carried in the pocket of the trousers (Phone).
\item \textbf{Smartwatch} fixed on the wrist of the right arm (Watch).
\item \textbf{Smartwatch} integrated into the helmet (Helmet).
\end{itemize}

A cyclist is identified with an ID, also called \textit{VRU (Vulnerable Road User)}, to make them distinguishable.
The captured data is labeled with the four movement primitives: ``waiting'' (class 0), ``starting'' (class 1), ``moving'' (class 2) and ``stopping'' (class 3).\\
To label the captured data by the four movement primitives, a wide-angle stereo-camera system is used to triangulate the cyclist's head. The triangulation area is depicted in Fig. \ref{crossing_camera}.
Both cameras have a sampling rate of 50 Hz.
The period of time when the cyclist enters the viewport and leaves the viewport of the camera is called a \textit{scene}.

\subsection{Preprocessing}
\label{cyclist_pre}
Each smart device captured data provided by three sensors, including an accelerator, a gyroscope as well as a rotation-vector sensor.
All supplied data are relative to the coordinate system with the device as the origin.
Since the devices are fixed on different positions on the cyclist's body, there are three different coordinate systems in total.
To boost the predictor's quality, all coordinate systems are transferred in a coordinate system relative to the cyclist \cite{BDS+18}.
For the process of segmentation, a sliding window approach is used as described in \cite{timeseriesbook}.
Every feature is calculated for different window lengths.
For feature extraction \textit{minimum}, \textit{maximum} as well as the \textit{energy} are chosen as calculation methods, see \cite{BBS14} and \cite{BDS+18} for further details.
For each device, 246 features are extracted, which results in 738 features in total.

\subsection{Ensemble Members}
\label{base learners_cyclists}
 The ensemble members are trained as pointed out in \cite{depping} using the \textit{data diversity} principle. Since the number of features is very high, a feature selection is necessary to reduce the dimension. A combination of filters and wrappers is used to decrease the number of features for all three smart devices.
After the feature selection, a \textit{XGBoost (Extreme Gradient Regression Trees)} \cite{xgboost} classifier is trained for each smart device.
The classifiers are optimized regarding the \textit{F1-score}.

\subsection{Ensemble Training}
\label{cyclist_csge}
The three pre-trained classifiers from Section \ref{base learners_cyclists} are used as the ensemble members for the \textit{XCSGE}. Altogether we have trained four different \textit{XCSGE} ensembles which will be explained later in detail.
For all experiments, the \textit{XCSGE} is optimized on the \textit{log loss} score. For the reason that the ensemble members predictions are class propabilities, the \textit{log loss} score is suitable.
We evaluated four different \textit{XCSGE} variants, described in more detail below:

\textbf{XCSGE with 4 PCA dimensions per smart device:}\hfill \\
As already pointed out, there are 246 features for all three smart devices.
After the fusion of the \textit{feature spaces} the dataset contains $3 \cdot 246 = 738$ features in total. The great amount of features increases the computational effort when using a \textit{k-nearest neighbor regressor} for the \textit{local weighting}. Therefore, we reduce the dimensionality. We applied a \textit{PCA} on the dataset. The reduced dataset contains four features per smart device, resulting in $3 \cdot 4 = 12$ features in total. We use the reduced dataset to train a \textit{k-nearest neighbor regressor}, which we then utilize for the \textit{local weighting} aspect.\\
\textbf{XCSGE with 50 PCA dimensions per smart device:}\hfill \\
To investigate the influence of the \textit{PCA} dimension, a \textit{PCA} is applied on the dataset, but this time to obtain a reduction to 50 features per smart device. The reduced feature set contains 50 features per smart device, resulting in $3 \cdot 50 = 150$ features in total. We use the reduced feature set to train a \textit{k-nearest neighbor regressor}, which we then utilize for the \textit{local weighting} aspect.\\
\textbf{XCSGE time-lagged:}\hfill \\
As a next approach, a \textit{XCSGE} with time-lagged ensemble members is trained.
Therefore, the predictions of all three smart devices are shifted back in time $(t=-14)$. This approach results in 14 ensemble members for each smart device. With a sampling rate of 0.02 seconds, a timespan of $14 \cdot 0.02s = 0.28s$ is covered.
Because of $14 \cdot 3 = 42$ ensemble members, the runtime of the \textit{XCSGE} is already relatively long. Therefore a PCA with four dimensions for each smart device is applied to shrink the \textit{feature space}.
 
\textbf{XCSGE with \textit{MLP regressor} for \textit{local weighting}:}\hfill \\
In this approach, we use all 738 features to train an \textit{MLP regressor}, which we use for the \textit{local weighting} model $M_{local}^{(j)}$, for details see Section \ref{sec:local}. We choose \textit{Rectifier} as the activation function and an architecture with 4 hidden layers and 100 neurons each.

\subsubsection{Analysis of local weighting models}
\label{cyclist:evaluation_local}

\begin{figure}[h!]
  \centering
  
    \includegraphics[scale=0.38]{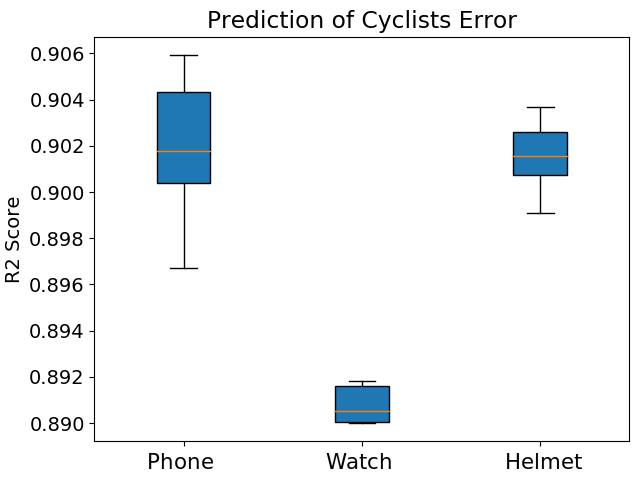}
\vspace{10px}
     \caption{\footnotesize R2-score of the \textit{local weighting} model $M_{local}^{(j)}$ using a \textit{MLP regressor}.}
    \label{error_local_cyclist}
\end{figure}

We have investigated the models, which are used for the estimation of the local error, in a six-fold cross-validation. The \textit{MLP regressor} was able to estimate the expected error best. The results can be seen in Figure \ref{error_local_cyclist}. For each ensemble member, it achieved a \textit{R2-score} of 0.9.

\subsection{Evaluation}
\label{cyclist_evaluation}
To evaluate the models, a six-fold cross-validation is executed. The results can be seen in Table \ref{los-loss-cyclist} and \ref{f1-cyclist}.
In each fold, the data is split in a training set (75\%) and in a test set (25\%).
The \textit{XCSGE} variations are trained on the training set.
Since the ensemble members are already pretrained \cite{depping}, there is no need for an extra training set.
The data is split by the \textit{VRU} to keep the structure of the scenes.
An evaluation of the four different \textit{XCSGE} ensembles is given below. The results can be seen in Tables \ref{los-loss-cyclist} and \ref{f1-cyclist}.

\textbf{XCSGE with 4 PCA dimensions per smart device:}\hfill \\
 The \textit{XCSGE} with 4 PCA dimensions increased its \textit{log loss} value at around 7\% compared to the best smart device model (Phone). When considering the \textit{F1-score}, the \textit{XCSGE} is as good as the best smart device model (Watch).\\

\textbf{XCSGE with 50 PCA dimensions per smart device:}\hfill \\
The \textit{XCSGE} with 50 PCA dimensions increased its \textit{log loss} value also at around 7 \%. When considering the \textit{F1-score} there is no significant improvement compared to the best base estimator (Watch).

\textbf{XCSGE time lagged:}\hfill \\
The \textit{time lagged XCSGE} increased the \textit{log loss} value at around 8\% compared to Phone. When considering the \textit{F1-score}, an improvement at around 0.6\% compared to the best smart devices models (Watch) can be seen.
 
\textbf{XCSGE with \textit{MLP regressor} for \textit{local weighting}:}\hfill \\
The \textit{XCSGE} increased the \textit{log loss} value at around 10 \% compared to Phone. When considering the \textit{F1-score}, an improvement at around 1.5\% compared to the best smart devices model (Watch) can be seen.
Furthermore, we evaluated the trained \textit{MLP regressor}, which is used for the \textit{local weighting}. At a three-fold cross-validation the \textit{MLP regressor} achieved a \textit{R2-score} of $0.95$, as well as a \textit{mean-squared-error} of $0.062$ per smart device. A further advantage over to the \textit{MLP regressor approach} lies in the comparatively much lower time required for training.

\begin{figure}[h!]
  \centering
  
    \includegraphics[scale=0.19]{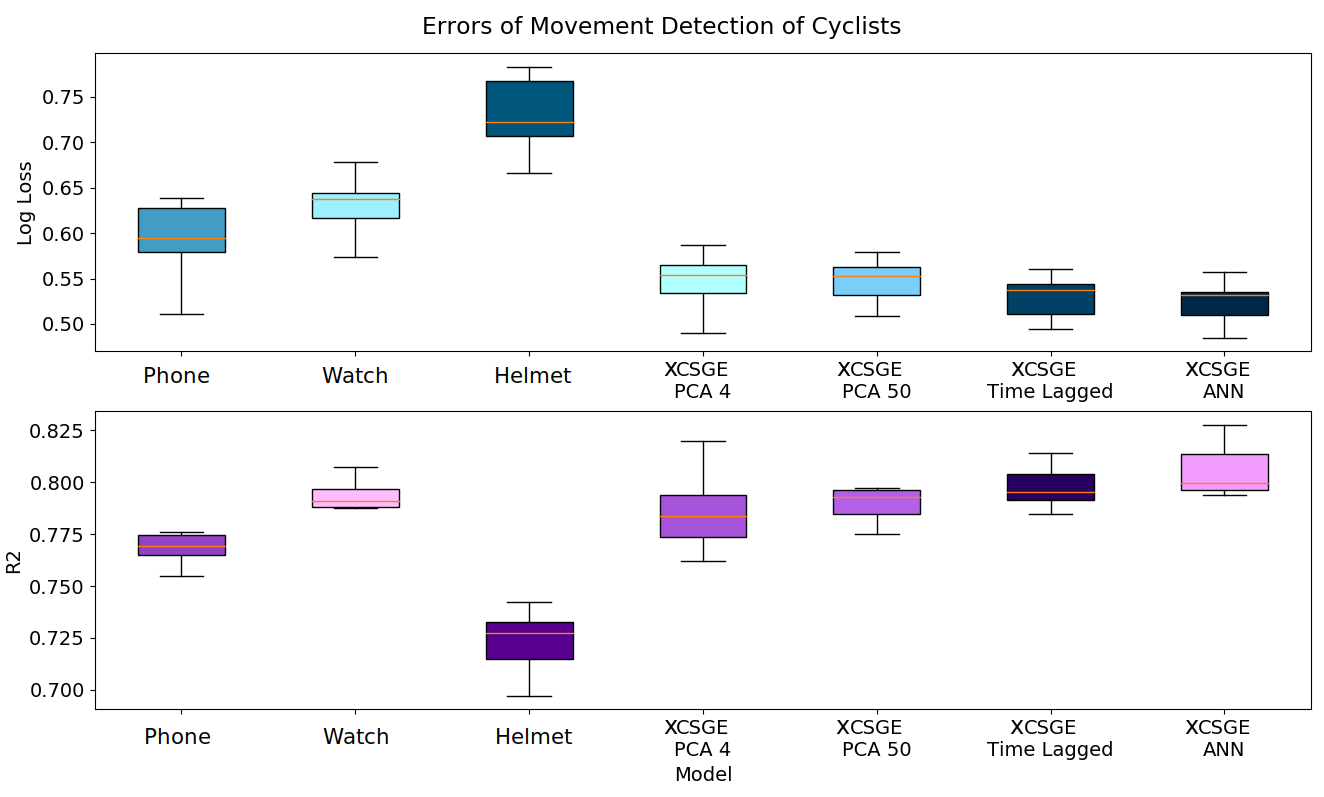}
    \vspace{0px}
     \caption{\footnotesize Box plot of \textit{log loss} and \textit{R2-score} of cyclists movement detection.}
    \label{box_wind}
\end{figure}

\resizebox{!}{.032\paperheight}{
\begin{tabular}{l|ccc|cc}
\toprule
 &  Phone &  Watch &  Helmet &  XCSGE (PCA Dim. 4) &  XCSGE (PCA Dim. 50)\\
\midrule
Mean         & 0.59211    & 0.63055    &       0.72944 & 0.54686    & 0.54754\\
Variance     & 0.00222     &    0.00125 &    0.00205 & 0.00114     & 0.00066 \\
Minimum     & 0.51139     &    0.57396 &    0.66575 & 0.49056    & 0.50922 \\
Maximum        & 0.63877     &    0.67849 &    0.78307 & 0.58715     & 0.57955 \\
Skill Score     & 18.83\%     & 13.56\%     & 0.0\%  & 25.03\%      & 24.94\% \\ 
\bottomrule
\end{tabular}
}

\vspace{15px}

\resizebox{!}{.04\paperheight}{
\begin{tabular}{l|cccc}
\toprule
 &  XCSGE (time lagged) & XCSGE (MLP regressor) \\
\midrule
Mean          & 0.529771 & \textbf{0.524286}      \\
Variance     & \textbf{0.000640} & 0.000672     &\\
Minimum      & 0.494970 & \textbf{0.48470}    \\
Maximum         & 0.559968 & \textbf{0.557599}    \\
Skill Score     &  26.72\% &  \textbf{28.14}\%     \\ 
\bottomrule
\vspace{10px}
\end{tabular}
}\captionof{table}{
\footnotesize Log loss of six-fold cross-validation. The best values for each criterium are highlighted bold.}
\label{los-loss-cyclist}

\vspace{10px}

\resizebox{!}{.032\paperheight}{
\begin{tabular}{l|ccc|cc}
&  Phone &  Watch &  Helmet &  XCSGE (PCA Dim. 4) &  XCSGE (PCA Dim. 50) \\
\midrule
Mean         & 0.77387    & 0.79147    &       0.72231 & 0.78613    & 0.79325    \\
Variance     & 0.00035     &    0.00013 &    0.00027 & 0.00041    & 0.00016      \\
Minimum     & 0.75498     &    0.77354 &    0.69707 & 0.76211    & 0.77949     \\
Maximum        & 0.80952     &    0.80757 &    0.74246 & 0.81978    & 0.81517     \\
Skill Score     & 6.66\%     & 8.87\%     & 0.0\%  & 8.12\%      & 8.94\%    \\ 
\bottomrule
\end{tabular}
}

\vspace{10px}

\resizebox{!}{.035\paperheight}{
\begin{tabular}{l|cccc}
 & XCSGE (time lagged) & XCSGE (MLP regressor) \\
\midrule
Mean          & 0.797745 & \textbf{0.806007} \\
Variance     & \textbf{0.000116}  & 0.000188\\
Minimum     & 0.784553 & \textbf{0.794117}\\
Maximum         & 0.814207 & \textbf{0.827823} \\
Skill Score     &  9.46\%  & \textbf{10.38}\% \\ 
\bottomrule
\end{tabular}
}
\vspace{10px}
\captionof{table}{\footnotesize F1-score of six-fold cross-validation. The best values for each criterium are highlighted bold.}
\label{f1-cyclist}

\vspace{20px}

\begin{figure}[!htb]
\minipage{0.5\textwidth}
  \includegraphics[width=\linewidth]{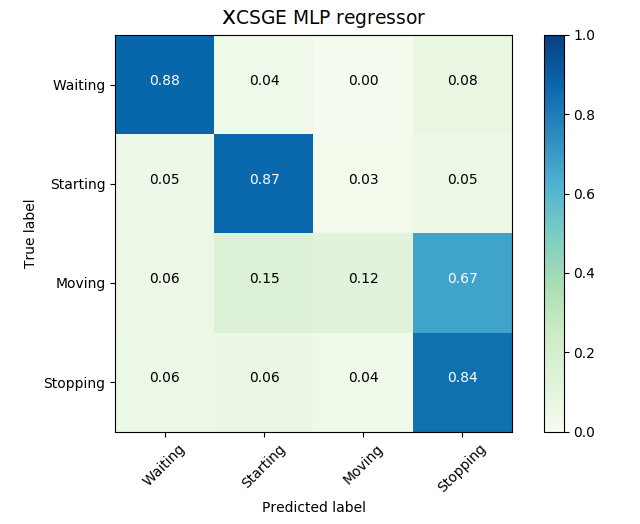}
\endminipage\hfill
\vspace{15px}
\caption{\footnotesize Confusions matrix of the XCSGE with a MLP regressor as local learning model.}\label{fig:conf_csge}
\end{figure}

Fig. \ref{fig:conf_csge} visualizes the confusions matrix of the best model (\textit{XCSGE MLP regressor}). We can see, that class ``waiting'' is recognized well by the classifier with a \textit{true-positive rate} (\textit{TP rate}) of $0.88$.\\
Class ``starting'' is also solidly classified by the predictor with a \textit{TP rate} of $0.87$.
It appears quite differently for class ``moving'' which is poorly recognized by the model and often classified as ``stopping''. Nearly all models classified most ``moving'' segments as ``stopping''. Nevertheless the ``stopping'' movement is classified well with a \textit{TP rate} of $0.84$ by the best \textit{XCSGE}.\\

\begin{figure}[h!]
  \centering
  
    \includegraphics[scale=0.4]{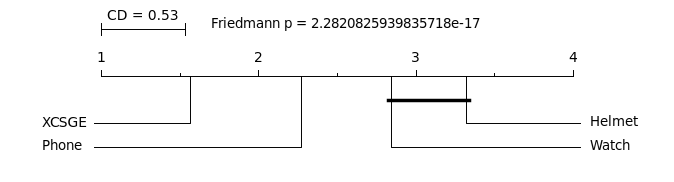}
\vspace{5px}
     \caption{\footnotesize Evaluation of ranked performance of the models using the \textit{Nemenyi post-hoc test}. As can be seen from the figure, the \textit{XCSGE} has the best ranked performance and is significantly better than the ensemble members.}
    \label{friedmann_cyclist}
\end{figure}

The ranked performance of the algorithms among all of the cyclists is furthermore analyzed using the \textit{Friedman test} in conjunction with the \textit{Nemenyi post-hoc test}. The results are given in Fig. \ref{friedmann_cyclist}. The \textit{Friedman p} value given in
the figure indicates that the ranks are significantly different, using a significance level of $\alpha=0.05$. As can be seen from the \textit{Nemenyi test}, the \textit{XCSGE} has a significantly better ranked performance in comparison
to ensemble members.

When considering the results of Table \ref{los-loss-cyclist} and \ref{f1-cyclist} it is conspicuous that all models have relatively good \textit{F1-} and \textit{log loss} scores compared to the  confusions matrices, while having big problems to identify ``moving'' and ``stopping'' correctly.
The distribution of the training data gives more insight into this problem. Most of the captured data is labeled as ``waiting'' (69.2 \%) and ``starting'' (14.1 \%).
This fact explains why the \textit{F1-} and \textit{log loss} score is relatively good compared to the confusions matrices.
The results allow some conclusions to be drawn. On the one hand, the results of the classifiers of the smartwatch on the wrist is significantly more reliable than those of the classifiers of the smartwatch integrated in the helmet  and the smartphone in the pocket. This is particularly noticeable in the motion primitive ``starting'', ``moving'' and ``stopping''. This is probably caused by the movement of the head of the cyclist to orientate himself in road traffic.
On the other hand, we could see that the selection of the machine learning algorithm $M_{local}^{(j)}$ to estimate the expected error for the \textit{local weighting} has a significant influence on the classification performance.

\section{Conclusion and Outlook}
\label{sec:outlook}

In this article, we have presented an extended version of the \textit{CSGE}.
Besides the possibility of using it for both regression and classification tasks, multivariate predictions are also introduced.
Furthermore, it is now possible to choose any machine learning algorithm for estimating the expected model error and therefore estimate the \textit{local weighting}.
Among the methodological aspects, one focus of the article is the evaluation and application of the \textit{XCSGE} to two challenging domains.
In both applications, the \textit{XCSGE} was able to improve the prediction quality. On the wind farm dataset, the \textit{XCSGE} outperforms the baseline by $11\%$.
When considering the solar power forecasting, the \textit{XCSGE} improves by $30\%$ compared to our baseline. In addition, the \textit{CSGE} outperformed the two \textit{Stacking} reference ensembles in terms of both \textit{R2-score} and \textit{RMSE}.
In the classification of motion primitive, the \textit{CSGE} achieved a skill score of $28.14\%$ outperforming all trained single models.\\
Although the \textit{XCSGE} has achieved significant improvements, there are some ways to achieve even better results. During the experiments, it was observed that the selection of features for \textit{local weighting} has a large influence on the prediction quality.
This could be improved by a feature selection.
In addition, we observe that the selection of the learning algorithm for the \textit{local weighting} has an equally large influence. The learning algorithm of the \textit{local weighting} affects not only the prediction quality but also the runtime of the \textit{XCSGE}, since the \textit{local weighting} has to be recalculated for each prediction.
In future work, we aim to predict probability distributions in the context of regression. Such an approach has already been discussed in \cite{gensler_sick_probabilistic}. Furthermore, the implemented possibility for multivariate prediction facilitates this approach.

\section{Acknowledgment}
\small{
This work results from the project DeCoInt$^2$, supported by the German Research Foundation (DFG) within the priority program SPP 1835: ``Kooperativ interagierende Automobile", grant numbers SI 674/11-1. This work results from the project project Prophesy (0324104A) funded by BMWi (German Federal Ministry for Economic Affairs and Energy).
}



\bibliographystyle{IEEEtran}
%

{\small
\bibliography{IEEEabrv,references}
}
\vskip -15mm

\appendix
\section{Anhang}
\captionsetup{list=no}

\resizebox{!}{.14\paperheight}{
\begin{tabular}{lcccc|ccc}
\toprule
Wind farm &   GBRT &    MLP &    SVR &  Ridge Regression &   CSGE &  MLP Stacking &  Linear Stacking \\
\midrule
0  & \cellcolor{a}  0.154408  & \cellcolor{g}  0.167766  & \cellcolor{e}  0.161638  & \cellcolor{g}   0.167491  & \cellcolor{c}  0.157962  & \cellcolor{a}          0.154193  & \cellcolor{a}\textbf{          0.152878 } \\
1  & \cellcolor{c}  0.188159  & \cellcolor{g}  0.199758  & \cellcolor{b}  0.183075  & \cellcolor{f}   0.194212  & \cellcolor{c}  0.185897  & \cellcolor{a}          0.181829  & \cellcolor{a}\textbf{          0.180173 } \\
2  & \cellcolor{a}\textbf{  0.116796 } & \cellcolor{g}  0.127257  & \cellcolor{f}  0.125700  & \cellcolor{g}   0.128722  & \cellcolor{b}  0.118826  & \cellcolor{a}          0.118186  & \cellcolor{b}          0.118813  \\
3  & \cellcolor{f}  0.180536  & \cellcolor{e}  0.172285  & \cellcolor{f}  0.175446  & \cellcolor{a}\textbf{   0.135534 } & \cellcolor{e}  0.170188  & \cellcolor{f}          0.176578  & \cellcolor{g}          0.189382  \\
4  & \cellcolor{a}  0.144537  & \cellcolor{c}  0.149138  & \cellcolor{d}  0.149743  & \cellcolor{g}   0.157768  & \cellcolor{a}\textbf{  0.143386 } & \cellcolor{a}          0.145039  & \cellcolor{b}          0.146082  \\
5  & \cellcolor{a}\textbf{  0.128751 } & \cellcolor{g}  0.139164  & \cellcolor{d}  0.134533  & \cellcolor{g}   0.138922  & \cellcolor{a}  0.129134  & \cellcolor{b}          0.131302  & \cellcolor{b}          0.131714  \\
6  & \cellcolor{f}  0.148872  & \cellcolor{f}  0.148168  & \cellcolor{f}  0.146514  & \cellcolor{a}\textbf{   0.125116 } & \cellcolor{f}  0.146794  & \cellcolor{g}          0.153208  & \cellcolor{g}          0.150376  \\
7  & \cellcolor{a}  0.153418  & \cellcolor{e}  0.160233  & \cellcolor{c}  0.156852  & \cellcolor{g}   0.165175  & \cellcolor{a}  0.152011  & \cellcolor{a}          0.153378  & \cellcolor{a}\textbf{          0.151502 } \\

8  & \cellcolor{a}  0.089079  & \cellcolor{c}  0.092628  & \cellcolor{c}  0.094991  & \cellcolor{g}   0.105708  & \cellcolor{a}\textbf{  0.087014 } & \cellcolor{a}          0.088029  & \cellcolor{a}          0.088621  \\
9  & \cellcolor{c}  0.102133  & \cellcolor{g}  0.106732  & \cellcolor{g}  0.107729  & \cellcolor{g}   0.107395  & \cellcolor{b}  0.101795  & \cellcolor{a}          0.100571  & \cellcolor{a}\textbf{          0.099450 } \\
10 & \cellcolor{a}  0.112926  & \cellcolor{b}  0.115092  & \cellcolor{d}  0.119257  & \cellcolor{g}   0.126735  & \cellcolor{a}\textbf{  0.111995 } & \cellcolor{a}          0.113562  & \cellcolor{a}          0.113357  \\
11 & \cellcolor{a}  0.109788  & \cellcolor{d}  0.117483  & \cellcolor{c}  0.115697  & \cellcolor{g}   0.127913  & \cellcolor{a}  0.109606  & \cellcolor{a}\textbf{          0.109211 } & \cellcolor{a}          0.110016  \\
12 & \cellcolor{a}\textbf{  0.135940 } & \cellcolor{g}  0.147448  & \cellcolor{e}  0.142687  & \cellcolor{f}   0.145309  & \cellcolor{a}  0.136747  & \cellcolor{b}          0.139067  & \cellcolor{b}          0.138245  \\
13 & \cellcolor{a}  0.170988  & \cellcolor{d}  0.185408  & \cellcolor{a}  0.171546  & \cellcolor{g}   0.208926  & \cellcolor{a}  0.170603  & \cellcolor{a}          0.168242  & \cellcolor{a}\textbf{          0.167550 } \\
14 & \cellcolor{a}\textbf{  0.162157 } & \cellcolor{b}  0.177160  & \cellcolor{a}  0.167641  & \cellcolor{g}   0.217813  & \cellcolor{a}  0.163709  & \cellcolor{a}          0.165717  & \cellcolor{a}          0.167208  \\
15 & \cellcolor{a}  0.165643  & \cellcolor{e}  0.177492  & \cellcolor{c}  0.170731  & \cellcolor{g}   0.187246  & \cellcolor{a}\textbf{  0.163837 } & \cellcolor{a}          0.166022  & \cellcolor{a}          0.165134  \\
16 & \cellcolor{a}  0.161203  & \cellcolor{d}  0.169510  & \cellcolor{c}  0.166153  & \cellcolor{g}   0.180876  & \cellcolor{a}  0.160845  & \cellcolor{a}          0.161574  & \cellcolor{a}\textbf{          0.160187 } \\
17 & \cellcolor{a}  0.151141  & \cellcolor{g}  0.162964  & \cellcolor{a}  0.152612  & \cellcolor{g}   0.163346  & \cellcolor{a}\textbf{  0.150836 } & \cellcolor{b}          0.153208  & \cellcolor{a}          0.152197  \\
18 & \cellcolor{a}  0.139423  & \cellcolor{e}  0.149743  & \cellcolor{d}  0.146741  & \cellcolor{g}   0.156079  & \cellcolor{a}  0.139466  & \cellcolor{a}          0.138333  & \cellcolor{a}\textbf{          0.138166 } \\
19 & \cellcolor{c}  0.191074  & \cellcolor{g}  0.197998  & \cellcolor{c}  0.190245  & \cellcolor{d}   0.192346  & \cellcolor{a}\textbf{  0.186730 } & \cellcolor{a}          0.188269  & \cellcolor{a}          0.187471  \\
20 & \cellcolor{a}  0.116591  & \cellcolor{c}  0.120052  & \cellcolor{d}  0.122104  & \cellcolor{g}   0.129244  & \cellcolor{a}\textbf{  0.114846 } & \cellcolor{a}          0.116563  & \cellcolor{a}          0.115663  \\
21 & \cellcolor{a}  0.153584  & \cellcolor{d}  0.160039  & \cellcolor{c}  0.158044  & \cellcolor{g}   0.167694  & \cellcolor{a}\textbf{  0.152604 } & \cellcolor{a}          0.152613  & \cellcolor{a}          0.152620  \\
... & ... & ... & ... & ... & ... & ... & .. \\
61 & \cellcolor{b}  0.167682  & \cellcolor{c}  0.172795  & \cellcolor{a}  0.165012  & \cellcolor{g}   0.186881  & \cellcolor{a}\textbf{  0.163428 } & \cellcolor{a}          0.164919  & \cellcolor{a}          0.163622  \\
62 & \cellcolor{b}  0.167056  & \cellcolor{c}  0.173117  & \cellcolor{a}  0.165741  & \cellcolor{g}   0.200315  & \cellcolor{a}  0.161767  & \cellcolor{a}          0.163315  & \cellcolor{a}\textbf{          0.161280 } \\
63 & \cellcolor{b}  0.153225  & \cellcolor{b}  0.154215  & \cellcolor{a}  0.148521  & \cellcolor{g}   0.174197  & \cellcolor{a}\textbf{  0.146803 } & \cellcolor{a}          0.147633  & \cellcolor{a}          0.148188  \\
64 & \cellcolor{b}  0.158646  & \cellcolor{g}  0.169243  & \cellcolor{d}  0.162357  & \cellcolor{e}   0.164706  & \cellcolor{b}  0.157285  & \cellcolor{a}          0.156562  & \cellcolor{a}\textbf{          0.155192 } \\
65 & \cellcolor{b}  0.135013  & \cellcolor{f}  0.143714  & \cellcolor{c}  0.137253  & \cellcolor{g}   0.145937  & \cellcolor{a}\textbf{  0.132856 } & \cellcolor{a}          0.133134  & \cellcolor{a}          0.133001  \\
66 & \cellcolor{c}  0.157357  & \cellcolor{f}  0.163898  & \cellcolor{c}  0.157624  & \cellcolor{g}   0.167720  & \cellcolor{a}  0.153742  & \cellcolor{a}\textbf{          0.152671 } & \cellcolor{a}          0.152942  \\
67 & \cellcolor{a}  0.207854  & \cellcolor{b}  0.217850  & \cellcolor{a}  0.204091  & \cellcolor{g}   0.263975  & \cellcolor{a}  0.204859  & \cellcolor{a}          0.200204  & \cellcolor{a}\textbf{          0.199410 } \\
68 & \cellcolor{b}  0.138188  & \cellcolor{d}  0.141684  & \cellcolor{c}  0.141313  & \cellcolor{g}   0.150069  & \cellcolor{a}\textbf{  0.135053 } & \cellcolor{a}          0.135342  & \cellcolor{a}          0.135711  \\
69 & \cellcolor{a}  0.211502  & \cellcolor{c}  0.218222  & \cellcolor{a}  0.212174  & \cellcolor{g}   0.242279  & \cellcolor{a}\textbf{  0.207674 } & \cellcolor{a}          0.211454  & \cellcolor{a}          0.210316  \\
\hline\hline \\
Mean & 0.13659 & 0.14075 & 0.14393 & 0.15163 & \textbf{0.13493} & 0.135642 & 0.1356257\\
Variance &	0.0013 &	\textbf{0.00119} &	0.00149 &	0.001729 &	0.00129 &	0.001285 &	0.00124\\
Minimum &	0.078383 &	\textbf{0.07040} &	0.07107 &	0.079998 &	0.071181 &	0.0711566 &	0.08276\\
Maximum	& 0.26646 &	0.26069 &	0.27779 &	0.30150 &	0.26088 &	0.261054 &	\textbf{0.25879}\\
Skill Score & 9.92\% & 7.18\% & 5.08 \% & 0.0 \% & \textbf{11.01\%} & 10.54\% & 10.55 \% \\ 

\bottomrule
\end{tabular}
}
\vspace{15px}
\captionof{table}{\footnotesize \textit{RMSE} of 10-fold cross-validation for each wind farm. Cell color are from green (good results) over orange (medium results) to red (worst results). Furthermore the best \textit{RMSE} score for each wind farm is highlighted bold.}
\label{appendix_wind_1}

\resizebox{!}{.14\paperheight}{
\begin{tabular}{lcccc|ccc}
\toprule
Wind farm &   GBRT &    MLP &    SVR &  Ridge Regression &   CSGE &  MLP Stacking &  Linear Stacking \\
\midrule
0  & \cellcolor{a}  0.764451  & \cellcolor{g}  0.721934  & \cellcolor{e}  0.741876  & \cellcolor{g}  0.722846  & \cellcolor{c}  0.753484  & \cellcolor{a}         0.765108  & \cellcolor{a}\textbf{         0.769098 } \\
1  & \cellcolor{c}  0.640584  & \cellcolor{g}  0.594908  & \cellcolor{a}  0.659747  & \cellcolor{e}  0.617089  & \cellcolor{b}  0.649174  & \cellcolor{a}         0.664360  & \cellcolor{a}\textbf{         0.670448 } \\
2  & \cellcolor{a}\textbf{  0.559313 } & \cellcolor{g}  0.476831  & \cellcolor{f}  0.489559  & \cellcolor{g}  0.464719  & \cellcolor{b}  0.543861  & \cellcolor{a}         0.548755  & \cellcolor{b}         0.543957  \\
3  & \cellcolor{f} -1.214009  & \cellcolor{e} -1.016265  & \cellcolor{e} -1.090918  & \cellcolor{a}\textbf{ -0.247803 } & \cellcolor{e} -0.967479  & \cellcolor{f}        -1.117996  & \cellcolor{g}        -1.436290  \\
4  & \cellcolor{a}  0.621127  & \cellcolor{c}  0.596623  & \cellcolor{d}  0.593347  & \cellcolor{g}  0.548589  & \cellcolor{a}\textbf{  0.627140 } & \cellcolor{a}         0.618490  & \cellcolor{b}         0.612984  \\
5  & \cellcolor{a}\textbf{  0.280263 } & \cellcolor{g}  0.159138  & \cellcolor{d}  0.214173  & \cellcolor{g}  0.162063  & \cellcolor{a}  0.275985  & \cellcolor{b}         0.251464  & \cellcolor{b}         0.246756  \\
6  & \cellcolor{f} -0.639366  & \cellcolor{f} -0.623894  & \cellcolor{f} -0.587848  & \cellcolor{a}\textbf{ -0.157914 } & \cellcolor{f} -0.593923  & \cellcolor{g}        -0.736243  & \cellcolor{g}        -0.672657  \\
7  & \cellcolor{a}  0.727443  & \cellcolor{e}  0.702691  & \cellcolor{c}  0.715103  & \cellcolor{g}  0.684070  & \cellcolor{a}  0.732419  & \cellcolor{a}         0.727586  & \cellcolor{a}\textbf{         0.734209 } \\
8  & \cellcolor{a}  0.789592  & \cellcolor{b}  0.772491  & \cellcolor{c}  0.760733  & \cellcolor{g}  0.703700  & \cellcolor{a}\textbf{  0.799231 } & \cellcolor{a}         0.794523  & \cellcolor{a}         0.791748  \\
9  & \cellcolor{c}  0.729704  & \cellcolor{g}  0.704809  & \cellcolor{g}  0.699272  & \cellcolor{g}  0.701130  & \cellcolor{b}  0.731489  & \cellcolor{a}         0.737906  & \cellcolor{a}\textbf{         0.743715 } \\
10 & \cellcolor{a}  0.750203  & \cellcolor{b}  0.740526  & \cellcolor{d}  0.721409  & \cellcolor{g}  0.685374  & \cellcolor{a}\textbf{  0.754303 } & \cellcolor{a}         0.747380  & \cellcolor{a}         0.748291  \\
11 & \cellcolor{a}  0.792793  & \cellcolor{c}  0.762727  & \cellcolor{c}  0.769886  & \cellcolor{g}  0.718727  & \cellcolor{a}  0.793479  & \cellcolor{a}\textbf{         0.794965 } & \cellcolor{a}         0.791930  \\
12 & \cellcolor{a}\textbf{  0.740582 } & \cellcolor{g}  0.694799  & \cellcolor{e}  0.714193  & \cellcolor{f}  0.703592  & \cellcolor{a}  0.737492  & \cellcolor{b}         0.728508  & \cellcolor{b}         0.731710  \\
13 & \cellcolor{a}  0.768628  & \cellcolor{c}  0.727957  & \cellcolor{a}  0.767114  & \cellcolor{g}  0.654564  & \cellcolor{a}  0.769668  & \cellcolor{a}         0.775998  & \cellcolor{a}\textbf{         0.777837 } \\
14 & \cellcolor{a}\textbf{  0.581002 } & \cellcolor{b}  0.499880  & \cellcolor{a}  0.552181  & \cellcolor{g}  0.244019  & \cellcolor{a}  0.572940  & \cellcolor{a}         0.562401  & \cellcolor{a}         0.554490  \\
15 & \cellcolor{a}  0.785676  & \cellcolor{d}  0.753917  & \cellcolor{b}  0.772308  & \cellcolor{g}  0.726127  & \cellcolor{a}\textbf{  0.790324 } & \cellcolor{a}         0.784695  & \cellcolor{a}         0.786991  \\
16 & \cellcolor{a}  0.755640  & \cellcolor{d}  0.729808  & \cellcolor{b}  0.740403  & \cellcolor{g}  0.692358  & \cellcolor{a}  0.756724  & \cellcolor{a}         0.754515  & \cellcolor{a}\textbf{         0.758710 } \\
17 & \cellcolor{a}  0.701320  & \cellcolor{g}  0.652761  & \cellcolor{a}  0.695475  & \cellcolor{g}  0.651134  & \cellcolor{a}\textbf{  0.702521 } & \cellcolor{b}         0.693094  & \cellcolor{a}         0.697131  \\
18 & \cellcolor{a}  0.719032  & \cellcolor{e}  0.675897  & \cellcolor{d}  0.688764  & \cellcolor{g}  0.647890  & \cellcolor{a}  0.718858  & \cellcolor{a}         0.723409  & \cellcolor{a}\textbf{         0.724077 } \\
19 & \cellcolor{c}  0.626200  & \cellcolor{g}  0.598621  & \cellcolor{c}  0.629438  & \cellcolor{d}  0.621208  & \cellcolor{a}\textbf{  0.643006 } & \cellcolor{a}         0.637095  & \cellcolor{a}         0.640168  \\
20 & \cellcolor{a}  0.719901  & \cellcolor{c}  0.703024  & \cellcolor{d}  0.692785  & \cellcolor{g}  0.655804  & \cellcolor{a}\textbf{  0.728220 } & \cellcolor{a}         0.720034  & \cellcolor{a}         0.724338  \\
21 & \cellcolor{a}  0.728310  & \cellcolor{d}  0.704992  & \cellcolor{c}  0.712302  & \cellcolor{g}  0.676093  & \cellcolor{a}\textbf{  0.731764 } & \cellcolor{a}         0.731734  & \cellcolor{a}         0.731711  \\
... & ... & ... & ... & ... & ... & ... & .. \\
61 & \cellcolor{b}  0.711448  & \cellcolor{c}  0.693580  & \cellcolor{a}  0.720564  & \cellcolor{g}  0.641586  & \cellcolor{a}\textbf{  0.725902 } & \cellcolor{a}         0.720879  & \cellcolor{a}         0.725252  \\
62 & \cellcolor{a}  0.665769  & \cellcolor{b}  0.641074  & \cellcolor{a}  0.671011  & \cellcolor{g}  0.519439  & \cellcolor{a}  0.686595  & \cellcolor{a}         0.680569  & \cellcolor{a}\textbf{         0.688482 } \\
63 & \cellcolor{b}  0.774146  & \cellcolor{b}  0.771219  & \cellcolor{a}  0.787800  & \cellcolor{g}  0.708092  & \cellcolor{a}\textbf{  0.792682 } & \cellcolor{a}         0.790332  & \cellcolor{a}         0.788752  \\
64 & \cellcolor{b}  0.561496  & \cellcolor{g}  0.500960  & \cellcolor{d}  0.540742  & \cellcolor{e}  0.527358  & \cellcolor{b}  0.568986  & \cellcolor{a}         0.572939  & \cellcolor{a}\textbf{         0.580384 } \\
65 & \cellcolor{b}  0.743854  & \cellcolor{f}  0.709776  & \cellcolor{c}  0.735286  & \cellcolor{g}  0.700730  & \cellcolor{a}\textbf{  0.751975 } & \cellcolor{a}         0.750934  & \cellcolor{a}         0.751434  \\
66 & \cellcolor{c}  0.715581  & \cellcolor{f}  0.691446  & \cellcolor{c}  0.714615  & \cellcolor{g}  0.676887  & \cellcolor{a}  0.728500  & \cellcolor{a}\textbf{         0.732268 } & \cellcolor{a}         0.731316  \\
67 & \cellcolor{a}  0.569236  & \cellcolor{b}  0.526809  & \cellcolor{a}  0.584692  & \cellcolor{g}  0.305222  & \cellcolor{a}  0.581563  & \cellcolor{a}         0.600363  & \cellcolor{a}\textbf{         0.603526 } \\
68 & \cellcolor{b}  0.689562  & \cellcolor{d}  0.673655  & \cellcolor{c}  0.675362  & \cellcolor{g}  0.633889  & \cellcolor{a}\textbf{  0.703491 } & \cellcolor{a}         0.702220  & \cellcolor{a}         0.700594  \\
69 & \cellcolor{a}  0.719318  & \cellcolor{c}  0.701197  & \cellcolor{a}  0.717531  & \cellcolor{g}  0.631687  & \cellcolor{a}\textbf{  0.729385 } & \cellcolor{a}         0.719446  & \cellcolor{a}         0.722457  \\
\bottomrule

\hline\hline \\
Mean & 0.66148	& 0.64301 &	0.63090 &	0.60955 &	\textbf{0.67195}	& 0.66523 &	0.66082 \\
Variance &	0.08610	& 0.07744	&0.07632	&\textbf{0.03427}	&0.07268	&0.08567	&0.10130\\
Minimum &	-1.21401&	-1.09092&	-1.01626&	\textbf{-0.24780}&	-0.96748&	-1.11800&	-1.43629\\
Maximum	& 0.83705	&0.81447	&0.81980	&0.76384	&0.84420	&0.84179	&\textbf{0.84651}\\
Skill Score & 8.52\% & 5.489\% & 3.502 \% & 0.0 \% & \textbf{10.237\%} & 9.13\% & 10.08 \% \\ 
\bottomrule
\end{tabular}
}
\vspace{15px}
\captionof{table}{\footnotesize R2-score of 10-fold cross-validation for each wind farm. Cell color are from green (good results) over orange (medium results) to red (worst results). Furthermore the best score for each wind farm is highlighted bold.
\label{appendix_wind_2}
} 

\newpage

\resizebox{!}{.10\paperheight}{

\begin{tabular}{lcccc|ccc}
\toprule
Solar farm &  GBRT &  MLP &   SVR &  Ridge Regression &  CSGE &  ANN Stacking &  Linear Stacking \\
\midrule
0   & \cellcolor{a}  0.043583  & \cellcolor{g}  0.075156  & \cellcolor{g}  0.073463  & \cellcolor{a}   0.044610  & \cellcolor{a}\textbf{  0.042633 } & \cellcolor{b}          0.049860  & \cellcolor{a}          0.045118  \\
1   & \cellcolor{a}  0.043583  & \cellcolor{g}  0.075156  & \cellcolor{g}  0.073463  & \cellcolor{a}   0.044610  & \cellcolor{a}\textbf{  0.042633 } & \cellcolor{b}          0.049860  & \cellcolor{a}          0.045118  \\
2   & \cellcolor{a}  0.053293  & \cellcolor{f}  0.073138  & \cellcolor{g}  0.080344  & \cellcolor{a}   0.056035  & \cellcolor{a}  0.052934  & \cellcolor{b}          0.057708  & \cellcolor{a}\textbf{          0.052723 } \\
3   & \cellcolor{a}  0.051670  & \cellcolor{g}  0.078083  & \cellcolor{g}  0.078452  & \cellcolor{a}   0.052923  & \cellcolor{a}\textbf{  0.050511 } & \cellcolor{b}          0.057120  & \cellcolor{a}          0.050994  \\
4   & \cellcolor{a}  0.061035  & \cellcolor{f}  0.080450  & \cellcolor{g}  0.085355  & \cellcolor{a}   0.063021  & \cellcolor{a}  0.059863  & \cellcolor{a}          0.062865  & \cellcolor{a}\textbf{          0.059548 } \\
5   & \cellcolor{a}  0.057092  & \cellcolor{g}  0.076015  & \cellcolor{g}  0.076011  & \cellcolor{a}   0.057436  & \cellcolor{a}\textbf{  0.055124 } & \cellcolor{b}          0.060809  & \cellcolor{a}          0.056373  \\
6   & \cellcolor{a}\textbf{  0.052334 } & \cellcolor{g}  0.079606  & \cellcolor{g}  0.077445  & \cellcolor{a}   0.055729  & \cellcolor{a}  0.052518  & \cellcolor{c}          0.060696  & \cellcolor{a}          0.054440  \\
7   & \cellcolor{a}  0.041172  & \cellcolor{e}  0.061036  & \cellcolor{g}  0.071999  & \cellcolor{a}   0.042758  & \cellcolor{a}\textbf{  0.040181 } & \cellcolor{b}          0.047515  & \cellcolor{a}          0.042379  \\
... & ... & ... & ... & ... & ... & ... & .. \\
103 & \cellcolor{a}  0.046076  & \cellcolor{g}  0.072989  & \cellcolor{g}  0.074602  & \cellcolor{a}   0.045060  & \cellcolor{a}\textbf{  0.044150 } & \cellcolor{c}          0.053633  & \cellcolor{a}          0.046273  \\
104 & \cellcolor{a}  0.063084  & \cellcolor{g}  0.087971  & \cellcolor{e}  0.076807  & \cellcolor{a}   0.064751  & \cellcolor{a}\textbf{  0.061786 } & \cellcolor{b}          0.069189  & \cellcolor{b}          0.066824  \\
105 & \cellcolor{a}  0.054209  & \cellcolor{g}  0.077862  & \cellcolor{g}  0.077068  & \cellcolor{a}   0.054335  & \cellcolor{a}\textbf{  0.052690 } & \cellcolor{c}          0.062651  & \cellcolor{a}          0.055359  \\
106 & \cellcolor{a}  0.068485  & \cellcolor{g}  0.086177  & \cellcolor{e}  0.078502  & \cellcolor{a}   0.068424  & \cellcolor{a}\textbf{  0.066064 } & \cellcolor{b}          0.070198  & \cellcolor{a}          0.067279  \\
107 & \cellcolor{a}  0.046341  & \cellcolor{f}  0.068006  & \cellcolor{g}  0.075914  & \cellcolor{a}   0.046776  & \cellcolor{a}\textbf{  0.045005 } & \cellcolor{b}          0.051235  & \cellcolor{a}          0.046548  \\
108 & \cellcolor{a}  0.049938  & \cellcolor{g}  0.072215  & \cellcolor{g}  0.074191  & \cellcolor{b}   0.053769  & \cellcolor{a}\textbf{  0.049177 } & \cellcolor{b}          0.056165  & \cellcolor{a}          0.052468  \\
109 & \cellcolor{a}  0.060261  & \cellcolor{g}  0.086665  & \cellcolor{e}  0.077289  & \cellcolor{b}   0.063456  & \cellcolor{a}\textbf{  0.059129 } & \cellcolor{b}          0.066767  & \cellcolor{a}          0.062633  \\
110 & \cellcolor{a}  0.055057  & \cellcolor{f}  0.073916  & \cellcolor{g}  0.077599  & \cellcolor{b}   0.058049  & \cellcolor{a}\textbf{  0.054575 } & \cellcolor{c}          0.064359  & \cellcolor{b}          0.059105  \\
111 & \cellcolor{a}  0.063440  & \cellcolor{g}  0.082062  & \cellcolor{f}  0.076722  & \cellcolor{b}   0.066341  & \cellcolor{a}\textbf{  0.062131 } & \cellcolor{c}          0.068641  & \cellcolor{a}          0.062850  \\
112 & \cellcolor{a}  0.064451  & \cellcolor{g}  0.083064  & \cellcolor{f}  0.079802  & \cellcolor{b}   0.068032  & \cellcolor{a}\textbf{  0.062703 } & \cellcolor{c}          0.069101  & \cellcolor{b}          0.065817  \\
113 & \cellcolor{b}  0.090967  & \cellcolor{g}  0.106739  & \cellcolor{d}  0.097555  & \cellcolor{b}   0.092172  & \cellcolor{a}\textbf{  0.086958 } & \cellcolor{c}          0.094661  & \cellcolor{b}          0.091249  \\
114 & \cellcolor{a}  0.047644  & \cellcolor{g}  0.076814  & \cellcolor{g}  0.075241  & \cellcolor{a}   0.047893  & \cellcolor{a}\textbf{  0.046460 } & \cellcolor{c}          0.055984  & \cellcolor{a}          0.049764  \\

\bottomrule

\hline\hline \\
Mean & 0.05955	& 0.07795 &	0.082801 &	0.061174 &	\textbf{0.05797}	& 0.065142 &	0.060682 \\
Variance &	$6.53E^{-5}$	& \textbf{$\boldsymbol{1.33E^{-5}}$}	& $4.56E^{-5}$	& $8.08E^{-5}$	& $6.051E^{-5}$	&$5.961E^{-5}$	&$6.69E^{-5}$\\
Minimum &	0.0411&	0.06909&	0.061035&	0.04177&	\textbf{0.0401811}& 0.04751 & 0.042378\\
Maximum	& 0.09096	&0.09755	&0.10673	&0.09217	&\textbf{0.08695}	&0.09466	&0.0912485\\
Skill Score & 28.08\% & 5.86\% & 0.00 \% & 26.12 \% & \textbf{30.00\%} & 21.33\% & 26.71 \% \\ 
\bottomrule
\end{tabular}
}
\vspace{5px}
\captionof{table}{\footnotesize\textit{RMSE} of 10-fold cross-validation for each solar farm. Cell color are from green (good results) over orange (medium results) to red (worst results). Furthermore the best \textit{RMSE} score for each solar farm is highlighted bold.\label{appendix_solar_1}
}

\resizebox{!}{.10\paperheight}{

\begin{tabular}{lcccc|ccc}
\toprule
Solar farm &  GBRT &  MLP &   SVR &  Ridge Regression &  CSGE &  ANN Stacking &  Linear Stacking \\
\midrule
0   & \cellcolor{a}  0.909879  & \cellcolor{g}  0.732009  & \cellcolor{g}  0.743945  & \cellcolor{a}  0.905581  & \cellcolor{a}\textbf{  0.913765 } & \cellcolor{b}         0.882048  & \cellcolor{a}         0.903419  \\
1   & \cellcolor{a}  0.909879  & \cellcolor{g}  0.732009  & \cellcolor{g}  0.743945  & \cellcolor{a}  0.905581  & \cellcolor{a}\textbf{  0.913765 } & \cellcolor{b}         0.882048  & \cellcolor{a}         0.903419  \\
2   & \cellcolor{a}  0.871502  & \cellcolor{e}  0.757988  & \cellcolor{g}  0.707950  & \cellcolor{a}  0.857941  & \cellcolor{a}  0.873230  & \cellcolor{b}         0.849332  & \cellcolor{a}\textbf{         0.874237 } \\
3   & \cellcolor{a}  0.870074  & \cellcolor{g}  0.703299  & \cellcolor{g}  0.700483  & \cellcolor{a}  0.863700  & \cellcolor{a}\textbf{  0.875841 } & \cellcolor{b}         0.841225  & \cellcolor{a}         0.873453  \\
4   & \cellcolor{a}  0.844117  & \cellcolor{f}  0.729169  & \cellcolor{g}  0.695140  & \cellcolor{a}  0.833803  & \cellcolor{a}  0.850044  & \cellcolor{a}         0.834626  & \cellcolor{a}\textbf{         0.851616 } \\
5   & \cellcolor{a}  0.884845  & \cellcolor{g}  0.795859  & \cellcolor{g}  0.795879  & \cellcolor{a}  0.883454  & \cellcolor{a}\textbf{  0.892645 } & \cellcolor{b}         0.869363  & \cellcolor{a}         0.887728  \\
6   & \cellcolor{a}\textbf{  0.894444 } & \cellcolor{g}  0.755762  & \cellcolor{g}  0.768845  & \cellcolor{a}  0.880304  & \cellcolor{a}  0.893699  & \cellcolor{b}         0.858016  & \cellcolor{a}         0.885777  \\
7   & \cellcolor{a}  0.916750  & \cellcolor{e}  0.817042  & \cellcolor{g}  0.745410  & \cellcolor{a}  0.910209  & \cellcolor{a}\textbf{  0.920708 } & \cellcolor{b}         0.889119  & \cellcolor{a}         0.911797  \\
... & ... & ... & ... & ... & ... & ... & .. \\
103 & \cellcolor{a}  0.917273  & \cellcolor{g}  0.792406  & \cellcolor{g}  0.783125  & \cellcolor{a}  0.920879  & \cellcolor{a}\textbf{  0.924042 } & \cellcolor{b}         0.887910  & \cellcolor{a}         0.916562  \\
104 & \cellcolor{a}  0.862044  & \cellcolor{g}  0.731726  & \cellcolor{d}  0.795494  & \cellcolor{a}  0.854660  & \cellcolor{a}\textbf{  0.867665 } & \cellcolor{b}         0.834051  & \cellcolor{b}         0.845203  \\
105 & \cellcolor{a}  0.900374  & \cellcolor{g}  0.794464  & \cellcolor{g}  0.798635  & \cellcolor{a}  0.899908  & \cellcolor{a}\textbf{  0.905880 } & \cellcolor{c}         0.866926  & \cellcolor{a}         0.896101  \\
106 & \cellcolor{a}  0.864132  & \cellcolor{g}  0.784866  & \cellcolor{e}  0.821481  & \cellcolor{a}  0.864376  & \cellcolor{a}\textbf{  0.873571 } & \cellcolor{b}         0.857252  & \cellcolor{a}         0.868875  \\
107 & \cellcolor{a}  0.907766  & \cellcolor{e}  0.801371  & \cellcolor{g}  0.752488  & \cellcolor{a}  0.906026  & \cellcolor{a}\textbf{  0.913009 } & \cellcolor{b}         0.887257  & \cellcolor{a}         0.906942  \\
108 & \cellcolor{a}  0.906950  & \cellcolor{g}  0.805416  & \cellcolor{g}  0.794619  & \cellcolor{b}  0.892124  & \cellcolor{a}\textbf{  0.909765 } & \cellcolor{b}         0.882299  & \cellcolor{a}         0.897284  \\
109 & \cellcolor{a}  0.892375  & \cellcolor{g}  0.777399  & \cellcolor{e}  0.822959  & \cellcolor{a}  0.880659  & \cellcolor{a}\textbf{  0.896382 } & \cellcolor{b}         0.867880  & \cellcolor{a}         0.883735  \\
110 & \cellcolor{a}  0.900891  & \cellcolor{f}  0.821363  & \cellcolor{g}  0.803118  & \cellcolor{a}  0.889824  & \cellcolor{a}\textbf{  0.902618 } & \cellcolor{c}         0.864570  & \cellcolor{b}         0.885781  \\
111 & \cellcolor{a}  0.871255  & \cellcolor{g}  0.784577  & \cellcolor{e}  0.811702  & \cellcolor{b}  0.859210  & \cellcolor{a}\textbf{  0.876512 } & \cellcolor{c}         0.849279  & \cellcolor{a}         0.873638  \\
112 & \cellcolor{a}  0.873676  & \cellcolor{g}  0.790177  & \cellcolor{f}  0.806332  & \cellcolor{b}  0.859248  & \cellcolor{a}\textbf{  0.880434 } & \cellcolor{b}         0.854788  & \cellcolor{a}         0.868262  \\
113 & \cellcolor{b}  0.833169  & \cellcolor{g}  0.770307  & \cellcolor{d}  0.808133  & \cellcolor{b}  0.828721  & \cellcolor{a}\textbf{  0.847552 } & \cellcolor{c}         0.819346  & \cellcolor{b}         0.832136  \\
114 & \cellcolor{a}  0.875118  & \cellcolor{g}  0.675396  & \cellcolor{g}  0.688554  & \cellcolor{a}  0.873813  & \cellcolor{a}\textbf{  0.881252 } & \cellcolor{b}         0.827575  & \cellcolor{a}         0.863757  \\
\bottomrule

\hline\hline \\
Mean & 0.88197	& 0.79503 &	0.77049 &	0.87510 &	\textbf{0.88816}	& 0.85877 &	0.87752 \\
Variance &	0.00057	& 0.00136	& 0.00139	& 0.00095	& \textbf{0.00049}	& 0.00059	& 0.00058\\
Minimum &	0.78043 &	0.68659 &	0.61026 &	0.68862 &	\textbf{0.79297} & 0.75592 & 0.76750\\
Maximum	& 0.93328	&0.90024	&0.87952	&0.92983	&\textbf{0.93929}	&92412	&0.93016\\
Skill Score & 14.47\% & 3.18\% & 0.00 \% & 13.58 \% & \textbf{15.27\%} & 11.46\% & 13.89 \% \\ 
\bottomrule
\end{tabular}
}
\vspace{5px}
\captionof{table}{\footnotesize\textit{R2-score} of 10-fold cross-validation for each solar farm. Cell color are from green (good results) over orange (medium results) to red (worst results). Furthermore the best \textit{RMSE} score for each solar farm is highlighted bold.}

\resizebox{!}{.06\paperheight}{ \label{appendix_solar_2}

\begin{tabular}{l|ccc|cc}
\toprule
{k-fold} &  DV1 &  DV9 &  DV10 &  CSGE (PCA Dim. 4) &  CSGE (PCA Dim. 50)\\
\midrule
0 & \cellcolor{c}  0.575802  & \cellcolor{c}  0.573963  & \cellcolor{g}  0.730817  & \cellcolor{c}  0.566619  & \cellcolor{b}  0.548545\\
1 & \cellcolor{e}  0.638133  & \cellcolor{d}  0.632127  & \cellcolor{g}  0.712950  & \cellcolor{a}  0.529792  & \cellcolor{a}  0.526733\\
2 & \cellcolor{c}  0.597778  & \cellcolor{f}  0.678493  & \cellcolor{g}  0.705238  & \cellcolor{b}  0.561072  & \cellcolor{b}  0.565145\\
3 & \cellcolor{d}  0.590590  & \cellcolor{f}  0.642852  & \cellcolor{g}  0.665752  & \cellcolor{a}  0.545934  & \cellcolor{b}  0.556041\\
4 & \cellcolor{a}  0.511387  & \cellcolor{c}  0.610768  & \cellcolor{g}  0.783068  & \cellcolor{a}  0.490565  & \cellcolor{a}  0.509222\\
5 & \cellcolor{c}  0.638767  & \cellcolor{c}  0.645117  & \cellcolor{g}  0.778830  & \cellcolor{a}  0.587148  & \cellcolor{a}  0.579553\\
\hline\hline\\
Mean 		& 0.59211	& 0.63055	&       0.72944 & 0.54686	& 0.54754\\
Variance 	& 0.00222 	&	0.00125 &	0.00205 & 0.00114 	& 0.00066 \\
Minimum 	& 0.51139 	&	0.57396 &	0.66575 & 0.49056	& 0.50922 \\
Maximum		& 0.63877 	&	0.67849 &	0.78307 & 0.58715 	& 0.57955 \\
Skill Score 	& 18.83\% 	& 13.56\% 	& 0.0\%  & 25.03\%  	& 24.94\% \\ 
\bottomrule
\end{tabular}
}

\vspace{20px}

\resizebox{!}{.06\paperheight}{
\begin{tabular}{l|cccc}
\toprule
{k-fold} &  CSGE (time lagged) & CSGE (MLP regressor) \\
\midrule
0  & \cellcolor{a}  0.494970  & \cellcolor{a}  0.503700  \\
1  & \cellcolor{a}  0.538282  & \cellcolor{a}  0.535390  \\
2  & \cellcolor{a}  0.545970  & \cellcolor{a}  0.535080  \\
3  & \cellcolor{a}  0.536332  & \cellcolor{a}  0.529100  \\
4  & \cellcolor{a}  0.503103  & \cellcolor{a}  0.484700 \\
5  & \cellcolor{a}  0.559968  & \cellcolor{a}  0.557599  \\
\hline\hline\\
Mean 		 & 0.529771 & 0.524286 	 \\
Variance 	& 0.000640 & 0.000672 	&\\
Minimum 	 & 0.494970 & 0.48470	\\
Maximum		 & 0.559968 & 0.557599	\\
Skill Score 	&  26.72\% &  28.14\%	 \\ 
\bottomrule
\end{tabular}
}\captionof{table}{\small Log-loss of 6-fold cross-validation. Cell color are from green (good results) over orange (medium results) to red (worst results). Furthermore the best values for each criterium are highlighted bold.}

\resizebox{!}{.06\paperheight}{
\begin{tabular}{l|ccc|cc}
{k-fold} &  DV1 &  DV9 &  DV10 &  CSGE (PCA Dim. 4) &  CSGE (PCA Dim. 50) \\
\midrule
0 & \cellcolor{d}  0.776240  & \cellcolor{a}  0.807575  & \cellcolor{g}  0.721974  & \cellcolor{b}  0.79533  & \cellcolor{b}  0.797329 \\
1 & \cellcolor{c}  0.770000  & \cellcolor{b}  0.788837  & \cellcolor{g}  0.727382  & \cellcolor{c}  0.77807  & \cellcolor{a}  0.792978\\
2 & \cellcolor{d}  0.763826  & \cellcolor{c}  0.773547  & \cellcolor{g}  0.734512  & \cellcolor{a}  0.78936  & \cellcolor{c}  0.774933\\
3 & \cellcolor{e}  0.768960  & \cellcolor{c}  0.793323  & \cellcolor{g}  0.742468  & \cellcolor{a}  0.81978  & \cellcolor{c}  0.792532\\
4 & \cellcolor{a}  0.809522  & \cellcolor{b}  0.798192  & \cellcolor{g}  0.697077  & \cellcolor{c}  0.77212  & \cellcolor{a}  0.815177\\
5 & \cellcolor{d}  0.754988  & \cellcolor{a}  0.787829  & \cellcolor{g}  0.710449  & \cellcolor{c}  0.76211  & \cellcolor{b}  0.782041\\
\bottomrule

\hline\hline \\
Mean 		& 0.77387	& 0.79147	&       0.72231 & 0.78613	& 0.79325	\\
Variance 	& 0.00035 	&	0.00013 &	0.00027 & 0.00041	& 0.00016 	 \\
Minimum 	& 0.75498 	&	0.77354 &	0.69707 & 0.76211	& 0.77949	 \\
Maximum		& 0.80952 	&	0.80757 &	0.74246 & 0.81978	& 0.81517 	\\
Skill Score 	& 6.66\% 	& 8.87\% 	& 0.0\%  & 8.12\%  	& 8.94\%	\\ 
\bottomrule
\end{tabular}
}

\vspace{10px}

\resizebox{!}{.08\paperheight}{
\begin{tabular}{l|cccc}
{k-fold}  & CSGE (time lagged) & CSGE (MLP regressor) \\
\midrule
\midrule
0 & \cellcolor{a}  0.806250  & \cellcolor{a}  0.818123  \\
1 & \cellcolor{a}  0.793690  & \cellcolor{a}  0.799542  \\
2 & \cellcolor{b}  0.784553  & \cellcolor{a}  0.795512  \\
3 & \cellcolor{c}  0.796924  & \cellcolor{b}  0.800092  \\
4 & \cellcolor{a}  0.814207  & \cellcolor{a}  0.827823  \\
5 & \cellcolor{a}  0.790849  & \cellcolor{a}  0.79411  \\
\hline\hline \\
Mean 		 & 0.797745 & 0.806007 \\
Variance 	& 0.000116  & 0.000188\\
Minimum 	& 0.784553 & 0.794117\\
Maximum		 & 0.814207 & 0.827823 \\
Skill Score 	&  9.46\%  & 10.38 \% \\ 
\bottomrule
\end{tabular}
}
\vspace{10px}
\captionof{table}{\small F1-score of 6-fold cross-validation. Cell color are from green (good results) over orange (medium results) to red (worst results). The best values for each criterium are highlighted bold.}

\end{document}